\documentclass[sw,crcready]{iosart2x}

\usepackage{float}
\usepackage{enumerate}
\usepackage{subcaption}
\usepackage{multicol}
\usepackage[linesnumbered,ruled,vlined]{algorithm2e}
\usepackage{graphicx}
\usepackage{amsmath} 
\usepackage{color, soul}
\usepackage{colortbl}
\usepackage[dvipsnames]{xcolor}

\definecolor{LightGray}{rgb}{0.97,0.97,0.97}
\usepackage{listings} 
\lstdefinelanguage{SPARQL}{
	basicstyle=\small\ttfamily,
	backgroundcolor=\color{LightGray},
	columns=fullflexible,
	breaklines=false,
	sensitive=true,
	frame=bt,
	aboveskip=1em,
	belowskip=1em,
	xleftmargin=.5em,
	xrightmargin=.5em,
	framexleftmargin=.5em,
	framextopmargin=.5em,
	framexbottommargin=.5em,
	framexrightmargin=.5em,
	tabsize = 2,
	showstringspaces=false,
	morecomment=[l][\color{gray}]{\#},       
	morecomment=[n][\color{blue}]{<http}{>}, 
	morestring=[b][\color{OliveGreen}]{\"},  
	keywordsprefix=?,
	classoffset=0,
	keywordstyle=\color{Sepia},
	morekeywords={@prefix},
	classoffset=1,
	keywordstyle=\color{Purple},
	morekeywords={rdf,rdfs,owl,xsd,purl,concept,property,entity},
	classoffset=2,
	keywordstyle=\color{MidnightBlue},
	morekeywords={
		SELECT,CONSTRUCT,DESCRIBE,ASK,WHERE,FROM,NAMED,PREFIX,BASE,OPTIONAL,
		FILTER,GRAPH,LIMIT,OFFSET,SERVICE,UNION,EXISTS,NOT,BINDINGS,MINUS,a
	}
}

\pubyear{0000}
\volume{0}
\firstpage{1}
\lastpage{1}

\begin{document}

\begin{frontmatter}

\title{Context-Aware Composition of Agent Policies by Markov Decision Process Entity Embeddings and Agent Ensembles}
\runtitle{Context-Aware Composition of Agent Policies}


\begin{aug}
\author[A]{\inits{N.}\fnms{Nicole} \snm{Merkle}\ead[label=e1]{nicole.merkle@kit.edu}%
\thanks{Corresponding author. \printead{e1}.}}
\author[A]{\inits{R.}\fnms{Ralf} \snm{Mikut}\ead[label=e2]{ralf.mikut@kit.edu}}
\address[A]{Institute for Automation and Applied Informatics (IAI), \orgname{Karlsruhe Institute of Technology (KIT)},
\cny{Germany}\printead[presep={\\}]{e1,e2}}

\end{aug}


\begin{abstract}
Computational agents support humans in many areas of life and are therefore found in heterogeneous contexts. This means that agents operate in rapidly changing environments and can be confronted with huge state and action spaces. In order to perform services and carry out activities satisfactorily, i.e. in a goal-oriented manner, agents require prior knowledge and therefore have to develop and pursue context-dependent policies. The problem here is that prescribing policies in advance is limited and inflexible, especially in dynamically changing environments. Moreover, the context (i.e. the external and internal state) of an agent determines its choice of actions. Since the environments in which agents operate can be stochastic and complex in terms of the number of states and feasible actions, activities are usually modelled in a simplified way by Markov decision processes so that, for example, agents with reinforcement learning are able to learn policies, i.e. state-action pairs, that help to capture the context and act accordingly to optimally perform activities. However, training policies for all possible contexts using reinforcement learning is time-consuming. A requirement and challenge for agents is to learn strategies quickly and respond immediately in cross-context environments and applications, e.g., the Internet, service robotics, cyber-physical systems. In this work, we propose a novel simulation-based approach that enables a) the representation of heterogeneous contexts through knowledge graphs and entity embeddings and b) the context-aware composition of policies on demand by ensembles of agents running in parallel. The evaluation we conducted with the "Virtual Home" dataset indicates that agents with a need to switch seamlessly between different contexts, e.g. in a home environment, can request on-demand composed policies that lead to the successful completion of context-appropriate activities without having to learn these policies in lengthy training steps and episodes, in contrast to agents that use reinforcement learning. The presented approach enables both context-aware and cross-context applicability of untrained computational agents. Furthermore, the source code of the approach as well as the generated data, i.e. the trained embeddings and the semantic representation of domestic activities, is open source and openly accessible on Github and Figshare.
\end{abstract}


\end{frontmatter}




\section{Introduction}
\noindent Computational agents operating in today's complex world have to make decisions by considering and executing various alternative actions and strategies that can complete an activity or task\footnote{The terms \emph{Task} and \emph{Activity} are considered synonymous in this work} and lead to a desired goal. However, depending on the complexity of an activity, many possible actions have to be weighed against each other according to the current context and utility in order to find an optimal strategy consisting of a sequence of possible and useful actions. Computational agents~\cite{Poole:2017} face this problem in complex and heterogeneous environments (e.g., the World Wide Web, domestic environments, industry, health-care) with many alternative courses of action that are supposed to generate immediate strategies depending on the environmental context~\cite{Verma:2018}. For instance, a service robot agent that is currently in the kitchen can perform many actions that represent different activities (e.g. washing dishes, putting dishes in the cupboard) that are generally part of the context of a kitchen~\cite{Burghart:2005,Kaiser:2018}. Actions can overlap, i.e. they belong to different activities and lead with a certain probability to different states. Furthermore, the service robot agent has to consider and perform different actions and activities when the contexts (i.e. states) and thus goals change~\cite{Weng:2021}. This means that the correct, i.e. most appropriate and suitable, sequence of actions need to be found by the agent so that the sequences of actions executed can lead to the desired execution and completion of an appropriate activity. \\

\noindent To be able to react adequately depending on the current state of the environment, many approaches~\cite{Mnih:2013,Mnih:2015,Silver:2016,Dennis:2020,Miralles-Pechu:2020,Garcia:2015,Fernandez:2010} use reinforcement learning (RL) to train and adapt policies. However, depending on the size of the state and action space, it can take a very long time (i.e. several thousand episodes + execution steps or hours and days) for an agent to learn appropriate policies that contribute to the successful completion of an activity. This in turn means that the on-the-fly integration, i.e. deployment and learning, of new policies for performing activities, is very time-consuming, as the required policies have to be trained before they can be applied in real environments. This also means that agents who need to operate seamlessly in different contexts would need to learn and know in advance all relevant policies for all intended activities in order to immediately perform an activity in a targeted manner, which is hardly feasible in multi-context environments where activities can be integrated afterwards. \\

\noindent To enable agents to operate immediately in individual, multi-context environments, we propose a simulation-based approach that builds on the idea that activities, i.e. states and actions of MDPs, can be represented by a knowledge graph and their spatial distribution in an n-dimensional entity embedding space whereas their context can be determined by their neighbourhood. In order to accelerate the composition of optimal, i.e. reward maximising,  policies, the simulation and parallel execution of potential actions by agents enables the simultaneous exploration of reward maximising policies. These considerations lead to the following hypothesis \emph{H}:
\begin{enumerate}[\bfseries H:]
    \item Semantic knowledge graph entities from Markov Decision Processes (MDP) which serve as basis for simulating environments and input to entity embedding vectors, help to constrain the state and action space of an agent, so that the selection and composition of useful and appropriate policies can be found much faster by an ensemble of agents than by an agent applying RL, i.e. deep-q learning, for training policies.
\end{enumerate}

\noindent Based on this hypothesis, this paper presents a platform that enables computational agents to request context-dependent composite policies (i.e. action sequences) ranked by relevance, i.e. obtained rewards.  \\

\noindent The prerequisite for training the required entity embedding vectors are datasets and semantic entities that reflect the agents' interactions with the environment. This type of data is usually obtained from observations that contain feedback to the agents. To facilitate the evaluation of our approach, we adopted the \emph{Virtual Home (VH)}\footnote{\url{http://virtual-home.org/tools/explore.html}} dataset, which contains 1563 descriptions (i.e. action sequences) of domestic activities, e.g. \emph{make coffee}, with 1973 atomic actions that can be performed by virtual agents (see ref.~\cite{puig:2018}). The dataset was required for building a KG of domestic activity entities and simulating environmental contexts and feedback, enabling the evaluated agents to learn or compose policies that reproduce domestic activities. \\

\noindent The research questions that will be answered in the context of this work, are:  \\

\begin{enumerate}[\bfseries RQ1:]
    \item Can the approach presented compose reward-maximising policies across contexts, i.e. across activities, that contribute to the successful completion of activities?  (Feasibility)
    \item Is the presented approach able to speed up policy delivery in terms of the required number of training steps and episodes compared to agents using RL, i.e. deep-q-learning neural networks (DQNN), for policy learning? (Learning velocity)
\end{enumerate}

\noindent This paper provides the following scientific contributions:
\begin{enumerate}[\bfseries C1:]
    \item A light-weight ontology and method that allows the automated generation of MDP knowledge graphs from activity datasets that reproduce MDPs.
    \item An approach for training MDP-related entity embeddings in order to constrain the different contexts and uncover semantic relatedness as well as similarities between MDP entities.
    \item A simulation function that utilises the MDP knowledge graphs in order to simulate the respective context of the agent and provide feedback about the appropriateness of a selected set of actions.
\end{enumerate}

\noindent The aim of this work is to enable computational agents, through the aforementioned contributions, to retrieve context-aware strategies on demand without having to learn reward-maximising strategies in lengthy and daunting training procedures. Furthermore, our approach is intended to enable agents to act across contexts and so that new contexts, i.e. activities, can be introduced into an environment at runtime. \\

\noindent The remainder of this paper is structured as follows. Sec.~\ref{sec:scenario} illustrates the intended application scenario and process flow of the proposed approach. Sec.~\ref{sec:foundations} provides the basic knowledge about MDPs, agent policies, value functions, and word embeddings that is required to understand the proposed approach. Sec.~\ref{sec:related_work} discusses related work in order to show the differences and benefits of the proposed approach compared to the related works. Sec.~\ref{sec:approach} uncovers the proposed approach that enables the context-aware composition of policies for cross-context activities. Sec.~\ref{sec:evaluation} illustrates the evaluation carried out and discusses the results obtained. Finally, Sec.~\ref{sec:conclusion} concludes this work and gives an outlook on future work.

\section{Application Scenario}
\label{sec:scenario}
\begin{figure*}[!htbp]
    \centering
  \includegraphics[width=\linewidth]{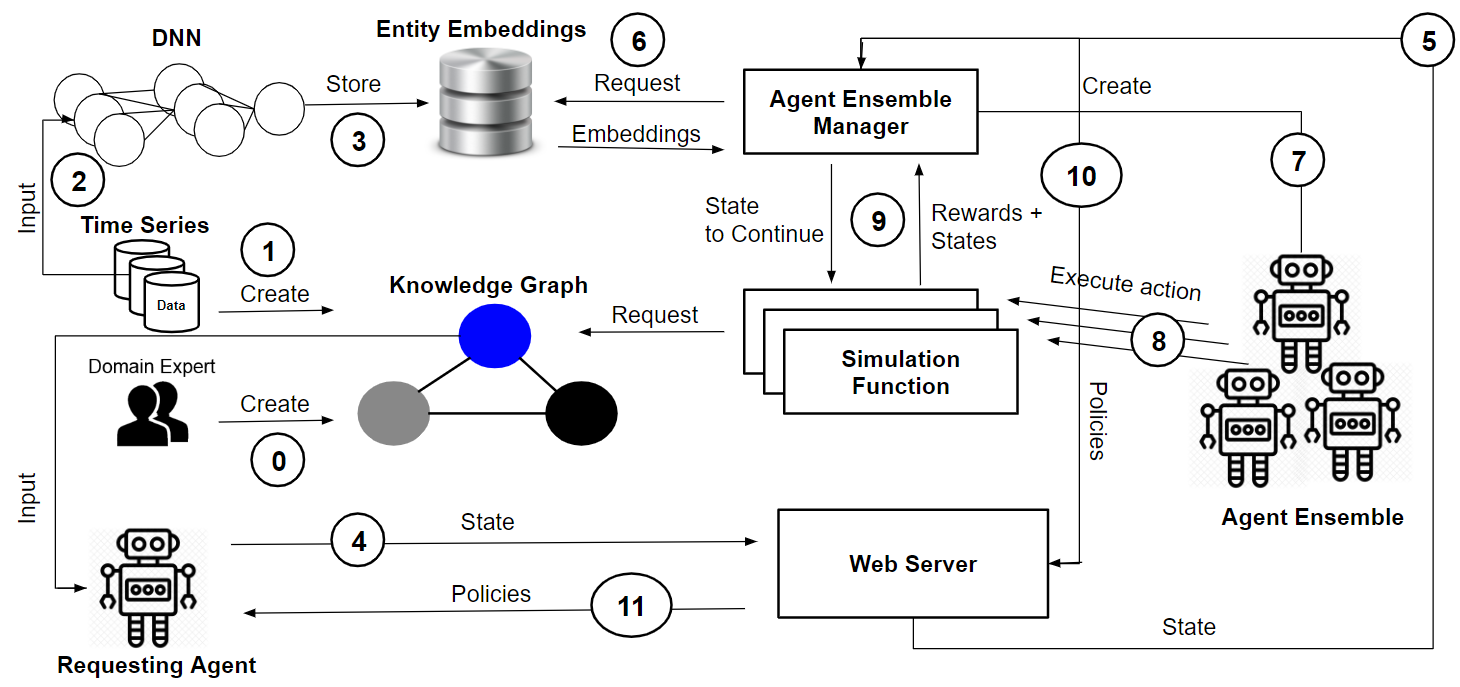}
  \caption{Contextual Policies Composition by Agent Ensembles and Entity Embeddings.}
  \label{fig:policies_composition}
\end{figure*}

\noindent The proposed approach aims at supporting computational agents in environments with heterogeneous contexts. For instance, service robots or household agents are envisaged that interact with unknown people and devices in unknown locations. In addition, agents might be transferred to other application fields and domains, e.g. to web or  cyber-physical systems. However, in the scope of this work, we consider and evaluate virtual domestic environments where rapid context changes are provided for an agent based on location changes, i.e. room changes, and prevailing devices and obstacles. In such a virtual environment, agents can perform all kinds of activities, such as making coffee or washing dishes, by performing sequences of atomic actions. In doing so, the sequence of actions requires  sometimes to be respected so that an activity can be carried out in a goal-oriented way. \\

\noindent An envisaged scenario would be that a household agent permanently observes the environment via its available sensors and recognises states on the basis of the received sensor data and a semantic model, i.e. knowledge graph. The agent now wants to find out which actions it can perform in its current state, or which actions the current state suggests. The task of the proposed system is to find possible action sequences for the agent based on the currently recognised state\footnote{Sometimes, however, the agent may observe a new, unknown state. In these uncertain situations, the system recognises that the state is unknown and suggests the agent to request feedback from the user and move to a safe position.} that best fit the agent's state or context. This means that the system implicitly recognises what would be the most obvious and profitable sequence of actions that the agent could perform in its current state. This requires that the system knows the possible activities and states in advance and can now use this knowledge to limit the agent's search space or context. The most suitable action sequences for the agent's current state are determined and reported back to the agent so that it can perform the most obvious activity. To do this, the approach simulates several activity scenarios using an ensemble of agents and determines the most profitable action sequence that is reward-maximising for the requesting agent and that it should perform in its current state. The agents of the ensemble can be viewed as individual processes or threats that simultaneously execute the found actions against the parallel running simulation functions. \\

\noindent Furthermore, recorded behaviours of the user or agent and feedback from the environment as well as the user can serve as useful cues for the agent to adapt its behaviour to the user's desired goals. For this reason, recorded log data can be used in the proposed approach to derive MDPs and introduce them as a new activity option in the system. The knowledge graph generated from this MDP then serves as the basis for the simulation of the activity and as input and extension for the embedded representation of the context. In this way, the agent's knowledge and context can be expanded over time to include alternative courses of action.\\

\noindent The proposed approach falls into the class of \emph{greedy} algorithms~\cite{Black:2020}, as it searches in each step for the optimal or most profitable action to find the best action sequence. Agents can also identify different alternative strategies to perform an activity. This is especially important when an agent cannot perform certain actions because it lacks the appropriate capabilities, i.e. actuators and devices. Alternative actions that are executable for the agent via available capabilities and actuators can thus be found for the agent. The agent then has the task of deciding on one of the action sequences proposed by the system by assessing which of the offered action sequences are executable for him. \\

\noindent Fig.~\ref{fig:policies_composition} illustrates the application scenario of the approach, which consists of 12 process steps. The general use case foresees that entity embedding vectors are continuously trained for agent activities, their associated states and atomic actions whenever a new semantic activity entity is either created by a domain expert \textbf{(step 0)} or automatically derived by supplied datasets \textbf{(step 1)} that reflect environmental sensor observations and allow the derivation of MDPs that represent activity entities. The aforementioned datasets are also used to train numerical embedding vectors representing the MDP entities contained in the dataset, i.e. activities, actions, states \textbf{(step 2)}. For this purpose, a vocabulary is created that assigns a unique numeric ID to all occurring entities. The IDs represent a numerical representation and serve as unique input values for training the embedding vectors. The resulting entity embeddings are stored together with the corresponding vocabulary in a database \textbf{(step 3)}. It is assumed that one or more agents are connected to the internet and located in heterogeneous environments who constantly are observing their environment with sensors. Based on these observations, an agent makes a request to the web server \textbf{(step 4)}, which forwards the request with the corresponding state information to a component called \emph{Agent Ensemble Generator} \textbf{(step 5)}. The \emph{Agent Ensemble Generator} loads pre-trained entity embedding vectors representing states and actions in an n-dimensional entity embedding vector space from a database \textbf{(step 6)}. \\ 

\noindent The \emph{Agent Ensemble Generator} determines the actions closest to the given state based on their trained and stored entity embedding vectors \textbf{(step 6)}. Two embedding vectors are considered to be close if their cosine distance is less than 1. In cases where no action is found between the distance range of 0 and 1, it can be assumed that an unknown or uncertain situation prevails where no action matches the given state. In such cases, the platform rejects the agent's request to avoid incorrect actions. In cases where actions are within the search radius of the current state, a set of the nearest actions is selected for execution. Then, the \emph{Agent Ensemble Generator} generates an ensemble of agent instances. The ensemble size depends on the set size of selected actions, since each agent of the ensemble executes exactly one action in each execution step \textbf{(step 7)}. The action commands in turn are executed by the respective ensemble agents in parallel threads that invoke a simulation function, which updates the states depending on the executed actions \textbf{(step 8)}. The simulation function performs for all agents the same simulation depending on the received action. In order to do so, it returns the updated state representation along with the resulting reward for each action-state pair to the \emph{Agent Ensemble Generator}, which then selects the action that yields the highest reward value \textbf{(step 9)}. The representation of activity, state and action concepts is defined by the platform's underlying ontology (see Sec.~\ref{sec:ontology}), which represents MDPs and is the basis for the related KG.  \\

\noindent The found policies, i.e. action sequence, are managed by the \emph{Agent Ensemble Generator} in a data structure, e.g. a table with action sequence lists (see Tab.~\ref{tab:policies}), until the executed activity is terminated through the simulation function by its goal and final state, which determines the end of the activity. The obtained and composed policies are ranked according to the reward values they received and sent back via the web server to the agent that initiated the request and provided the state information \textbf{(step 10, 11)}.  \\

\begin{table}[!htb]
	\centering
	\caption{Data Structure with the composed actions resp. policies. The rank determines the quality of the actions based on the reward values obtained.}
	\label{tab:policies}
	\resizebox{\textwidth}{!}{%
		\begin{tabular}{|c|cccc|}
			\hline
			\textbf{Rank (m)} & \multicolumn{4}{c|}{\textbf{Sequence (n)}}                                      \\ \hline
			1             & \multicolumn{1}{c|}{Action$_{1,1}$} & \multicolumn{1}{c|}{Action$_{1,2}$} & \multicolumn{1}{c|}{$\cdots$} & Action$_{1,n}$ \\ \hline
			2             & \multicolumn{1}{c|}{Action$_{2,1}$} & \multicolumn{1}{c|}{Action$_{2,2}$} & \multicolumn{1}{c|}{$\cdots$} & Action$_{2,n}$ \\ \hline
			$\cdots$             & \multicolumn{1}{c|}{$\cdots$} & \multicolumn{1}{c|}{$\cdots$} & \multicolumn{1}{c|}{$\cdots$} & $\cdots$ \\ \hline
			m             & \multicolumn{1}{c|}{Action$_{m,1}$} & \multicolumn{1}{c|}{Action$_{m,2}$} & \multicolumn{1}{c|}{$\cdots$} & Action$_{m,n}$ \\ \hline
		\end{tabular}%
	}
\end{table}

\noindent In this way, agents in heterogeneous environments can request policies from the web server based on their observed states. This saves the agents time-consuming learning processes, as the required knowledge is already implicitly encoded in activity-specific entity embedding vectors representing corresponding contexts. \\

\noindent In the used MDP model representation, semantic activity entities encapsulate states and actions that have an implicit semantic relationship, since an activity may consist of different states and require different actions that affect the observed states. These activity entities are accessible through the corresponding KG. \\

\noindent An arbitrary simulation function, adhering to the platform's ontology, is thus able to simulate activities based on the provided activity entities. The intended purpose of the simulation function is to update the states based on the actions performed and to provide feedback on the usefulness of the action performed. This feedback is represented by reward values, where a reward value less than 0 indicates undesirable or incorrect actions that should be avoided. \\

\noindent As mentioned earlier, a simulation function can also use activity entities extracted and generated from real operational datasets. In these cases, the simulation function can incorporate different MDP contexts, allowing for a wider range of individual contexts that reflect different environments of the agents.

\section{Foundations}
\label{sec:foundations}
\noindent This section lays the technical and formal foundations for understanding RL on the one hand and the approach of this paper on the other. For instance, the activities of agents are modelled and represented by MDPs in both RL and our approach. Moreover, strategies learned through value functions are expressed through policies. These terms require clarification. First, MDPs are introduced, then a definition of policies and value functions is given, and finally entity embeddings, which are one of the main building blocks of our approach, are discussed. For the description of \emph{finite MDPs, policies and value functions} in this section, we adopt the definitions from the RL book~\cite{Sutton:1998} by Sutton and Barto.

\subsection{Markov Decision Process} 
\noindent According to the considered literature, a \emph{finite} MDP is a mathematical formalization that allows RL agents to make sequential, goal-oriented decisions that do not only impact immediate rewards but also subsequent states and future (i.e. delayed) rewards. MDPs are defined through a tuple $(S, A, R)$ of finite sets, where \emph{S} is the finite set of process states, \emph{A} is the finite set of possible actions that can be performed by an agent and \emph{R} is a finite set of reward values. MDPs allow agents the learning of policies through their interactions with their external environment. Thereby, an agent has the objective to maximise through its performed actions and accumulated reward values, its internally maintained \emph{value} function. Each action $A_t \in A$ is selected and performed in discrete time steps $t = 0,1,2,3...n$ based on the perceived state $S_t \in S$. After executing an action $A_t$, the agent obtains a numerical reward signal $R_{t+1} \in R \subset \mathbb{R}$ and a subsequent state $S_{t+1}$. Thus, a sequence or \emph{trajectory} of a MDP is defined in~\cite{Sutton:1998} as follows: $S_0, A_0, R_1, S_1, A_1, R_2, S_2, A_2, R_3,...$.
The probability distribution $p$ of states and rewards is determined by their preceding states and conducted actions. Function~\ref{eq:prob} formalises this fact:
\begin{align}
\label{eq:prob}
    \mathsf{p(s',r | s, a)} \stackrel{.}{=} Pr\{S_t = s', R_t = r | S_{t-1} = s, A_{t-1} = a\}.
\end{align}
The function $p$ specifies a conditional probability and counts for all $s, s' \in S$, $r \in R$, $a \in A(s)$ and expresses the \emph{dynamics} of a MDP, defined through:
\begin{align} 
\label{eq:p_func}
p:S x R x S x A \rightarrow [0,1].
\end{align}
This also indicates the \emph{Markov property} of a state, since only the previous state and the action performed determine the probability of the next state $S_{t+1}$ and the corresponding reward $R_{t+1}$. Thus, it is not necessary to consider all previous states, but only the current state, to determine the transition probabilities of S and R. There is much more to say about MDPs, but the information outlined may suffice as a basis for this paper.

\subsection{Policies and Value Functions}
\noindent \emph{Policies} are closely related to MDPs and value functions. A \emph{value function} indicates how useful and reward-maximising a state or state-action pair is for the agent. Basically, the value function is maintained and used by agents to calculate expected future rewards. A value function requires always to be considered in relation to a policy where a policy represents a probability function $\pi(a|s)$ indicating the probability $a \in A(s)$ of an action $a = A_t \in A$ to be performed in state $s = S_t \in S$. According to Barto et al., a value function, computing the \emph{expected reward} $\mathbb{E}[.]$, w.r.t. a policy $\pi$, is defined as follows, where $\gamma$ is the \emph{discount factor} that controls how strong an immediate reward is affecting the value of the value function:
\begin{align}
    \label{eq:value_function}
    \mathsf{v}_\pi (s) \stackrel{.}{=} \mathbb{E}_\pi[G_t|S_t=s] = \mathbb{E}_\pi \left[\sum_{k=0}^{\infty} \gamma^k R_{t+k+1}\Bigg|S_t = s\right], \text{for all}\; s \in S.
\end{align}
The RL book makes a distinction between a \emph{state-value} function $v_\pi$ and an \emph{action-value} function $q_\pi$ that is defined in function~\ref{eq:act_val}:
\begin{align}
    \label{eq:act_val}
    \mathsf{q}_\pi(s|a) \stackrel{.}{=} \mathbb{E}_\pi[G_t|S_t = s, A_t = a] = \mathbb{E}_\pi\left[\sum_{k=0}^{\infty}\gamma^k R_{t+k+1}\Bigg|S_t = s, A_t = a \right], \text{for all}\; s \in S, \text{for all}\; a \in A.
\end{align}

\noindent The difference between the two value functions is that the latter takes into account not only the state but also the performed action. It is worthwhile to mention that the value functions are updated repeatedly for any number of episodes until the value functions converge. A training episode begins with an initial state and ends when the final state, i.e. goal state, is reached.
 
\subsection{Word Embeddings}
\noindent Word embeddings are used in particular in the field of natural language processing (NLP) and can be created by various methods (e.g. \emph{skip-gram}~\cite{NIPS:2013} and \emph{continuous bag of words (CBOW)}~\cite{Mikolov:2013}) to encode words numerically, since only numerical representations of categorical features (e.g. words) can be processed, especially in machine learning. Increasingly, word embeddings are used in combination with KGs. In this paper, word embeddings are considered as numerical (i.e. real-valued) representations of \emph{states, actions} and \emph{activities} in an n-dimensional vector space. An important aspect of word embeddings is that they capture the context or semantic relatedness and similarity of words or entities by their distribution and distance from each other in a defined vector space. For example, Allen and Hospedales have shown in their publication~\cite{Allen:2019} that semantic relations, e.g. analogies, can be revealed by simple mathematical operations (e.g. subtraction, addition) on word embeddings. 
The discussed properties of word embeddings are useful for our approach, as the intended goal is to capture the semantic relations of agents' activities and contexts in order to constrain the state and action space to speed up policy composition.

\section{Related Work}
\label{sec:related_work} 
\noindent RL is applied in many application fields concerned with computational agents, MDPs and decision making~\cite{Dohmatob:2020,KunLu:2021}. The drawback of RL is however, that policies are trained for specific activities and contexts and thus are not applicable across contexts and domains. Moreover, a disadvantage of agents that implement RL algorithms is that they usually require long training procedures in order to find reward maximising policies, especially in cases where the state and action space is huge. The following related works apply RL in different domains, however, it turns out that they are specific to different contexts since they are trained for specific activities in appropriate application domains. Furthermore, the fact that most of them apply RL, introduces the problem, that policies have to trained in advance in long training procedures, while the integration of new contexts and activities seems to be not addressed by the considered related works. \\

\noindent For instance, Liu et al. propose a data-driven deep-q-RL framework for learning policies of sequential treatment plans for patients~\cite{Liu:2017}. The approach combines supervised learning and RL for composing sequences of dynamic treatment plans. Similar to our approach, their objective is to reduce the state and action space and derive depending on the patient's context (i.e. patient data, vital parameters) suitable treatment plans. However, this approach seems only to be applicable to treatment plans in the medical context. \\

\noindent Agarwal et al. propose a \emph{"policy similarity metric (PSM)"} for \emph{"representation learning"} to enable better generalisation of behaviour to unknown agent environments. To achieve this, their approach \emph{"measures the behavioural similarity of states across different tasks"}~\cite{Agarwal:2021}. While the related approach also uses embeddings to encode states, our approach considers not only states but also actions and activities and aims at speeding up the process of finding context-appropriate strategies without the need to apply RL. This means that our approach omits RL whereas it ensures that agents receive assembled strategies that they can apply immediately when required and depending on their context.  \\

\noindent Da Silva et al. investigate \emph{inter-agent teaching} respectively \emph{transfer learning} that is applied for increasing the learning speed of RL agents with respect to policies~\cite{DaSilva:2020}. However, they point out that considered teaching approaches between agents are restricted regarding their ability to generalise to different domains and application scenarios. \\

\noindent Xian et al. propose \emph{Policy Guided Path Reasoning} that uses KGs and RL with a \emph{soft reward strategy} to reason and explain reasoning and recommendation paths~\cite{Xian:2019}. Thus, the focus of this work seems to be the explainability of the policy paths trained and not on the on-demand provision of context-dependent policies. \\

\noindent Zhao et al. present a method for making explainable path recommendations in a KG by means of a model named (ADversarial Actor Critic (ADAC)) model that combines RL with imitation learning~\cite{Zhao:2020}. Using imitation learning, implies that imitations are prevalent in advance that can be fed into the training procedure. \\

\noindent Kanervisto et al. discuss how manipulating (e.g. remove or categorise actions) the action space in RL problems, can influence positively the performance of RL applications~\cite{Kanervisto:2020}. However, the related approach presupposes that there is a knowledge of how the action space can be manipulated. The manipulation might require human intervention. In contrast to the related approach, our approach allows the automated restriction of the action space by entity embeddings and the prediction of the reward-maximising actions by activity datasets. \\

\noindent The approach of Hanawal et al. uses state-action samples representing intended policies in order to learn generalisable policies by means of \emph{$L_1$ regularised logistic regression}~\cite{Hanawal:2019}. \\

\noindent Chatzis et al. proposes a \emph{dynamic non-parametric Bayesian model} that addresses partially observable and dynamically changing MDPs~\cite{Chatzis:2014}, while Pandey et al. present a \emph{hybrid planning approach} that "combines deterministic planning and MDP planning for generating policy adaptation plans"~\cite{Pandey:2016}.  \\

\noindent Wang et al. suggest a \emph{multi-agent reinforcement learning approach} that allows the dynamic composition of services~ \cite{Wang:2016}. \\

\noindent The approach in Runck et al.'s paper~\cite{Runck:2019} utilises word embeddings to generate agent-based models for decision making that are derived from natural language descriptions of human behaviour.  \\

\noindent In contrast to the related work considered, we omit RL completely, as our addressed requirement is to provide policies on demand, as quickly as possible, and across contexts. RL approaches, however, require time-consuming (off-line) training of policies and are usually limited to activity contexts presented and taught in either real or simulated environments. However, we want agents to be able to perform actions in changing contexts without the requirement to learn policies (in advance). Thus, the proposed approach aims at avoiding the aforementioned drawbacks of RL. As some of the related work, we also use MDPs, knowledge graphs and embeddings for modelling and simulating activities. However, the focus of our work is the on-the-fly retrieval of polices across heterogeneous contexts and environments in order to answer on-demand requests from computational agents.

\section{Approach}
\label{sec:approach}
\noindent The approach section is divided in seven parts. First, we outline the interplay of the proposed architecture's software components. Second, we present the ontology that lays the foundation for creating MDP entities. Third, we outline how MDPs can be derived from activity log datasets. Fourth, we discuss how MDP entity embeddings from the ontology concepts, i.e. activity, state and action entities can be trained in order to semantically represent them in an embedding space that captures the different contexts. Fifth, we outline the simulation function that utilises the generated knowledge graph and serves for the ensemble agents as environment and resource of feedback. Sixth, we discuss the algorithm that implements the policies composition by agent ensembles. Lastly, we outline the preconditions and limitations of the presented approach.

\subsection{Interaction of the Software Components}
\label{sec:interplay}
\noindent The general operation and interaction of the software components is as follows: MDP knowledge graphs are either created by domain experts or derived from recorded time series datasets. Thus, the ontology provides the concepts and properties for building the knowledge graphs. The MDP knowledge graph describes what the state - action - state transitions look like. Based on these state transitions, the input to the deep neural network that encodes the embeddings is built. The embedding vectors are a numerical representation of the knowledge represented in the knowledge graph since they are trained based on the entities of the MDP knowledge graph and their relationships to each other. By means of the embedding vectors, the \emph{EnsembleAgentGenerator} can determine the most appropriate actions for a given state. The embeddings enable a fast search for spatially related actions, since we can apply similarity metrics to find related, i.e., nearby, actions and states. In addition, the knowledge graph provides the simulation functions with knowledge needed to simulate the states and evaluate the actions performed.

\subsection{The Concepts and Properties of the MDP Ontology}
\label{sec:ontology}
\noindent The proposed approach requires an ontology that defines the necessary concepts and properties to build and extend a knowledge graph based on activity datasets that represent MDPs. For this reason, we have devised a light-weight ontology that captures properties of MDPs already outlined in Sec.~\ref{sec:foundations}. Based on the developed ontology, interrelated knowledge graph entities are generated, which serve on the one hand as input for the encoding of entity embeddings and on the other hand as blueprint for the simulation function described in Sec.~\ref{sec:simulation}. It is expected that the simulation function needs to evaluate states, manage them based on the actions performed in each execution step, and report back to the agent the goodness, i.e. utility, of the action performed. This requires the simulation function to know what effects an action causes depending on the current state. In addition, there are rewards for executed actions and obtained states as well as transition probabilities that make the state changes stochastic. Taking all these aspects into account, all concepts and properties of the proposed ontology are discussed in detail below. The corresponding tables show the properties of the concepts, the object or value ranges, the cardinality and the requirement of the properties within a concept. The serialization formats we use for the ontology are JSON-LD\footnote{\url{https://json-ld.org/}} and Turtle\footnote{\url{https://www.w3.org/TR/turtle/}} RDF. Moreover, SPARQL\footnote{\url{https://www.w3.org/TR/sparql11-query/}} queries are performed for querying the available knowledge graphs.\\

\noindent The concept \emph{Activity} represents activities, i.e. MDPs that can be performed by agents. It therefore refers to all possible observations, states and actions within the activity. Since an activity can be performed by one or more agents, i.e. actors, the number of agents involved in the activity must be specified to determine when a new state change has occurred. Only when all agents have performed their action is a new state determined. Communication between agents and the simulation function can be asynchronous or synchronous. This is particularly relevant if several agents are involved in the activity and it must be determined whether the simulation function should react to the agents asynchronously or synchronously. Sometimes activities require adherence to a fixed sequence of actions. In these cases, the activities are sequential, i.e. the property \emph{:isSequential} is set to \emph{true}. In other cases, no strict order of actions is required to perform an activity. In these cases, the property \emph{:isSequential} is set to \emph{false}. \\ 

\begin{table}[!htb]
		\centering
		\caption{Activity concept and its properties.}
		\label{tab:task_class}
		\resizebox{\textwidth}{!}{%
	\begin{tabular}{|c|c|c|c|}
		\hline
		\multicolumn{4}{|c|}{\textbf{Concept:Activity}}                                                            \\ \hline
		\textbf{Properties} & \textbf{Range}               & \textbf{Cardinality} & \textbf{Mandatory} \\ \hline
		:hasState                  & :State                         & 1..*                & Yes    \\ \hline
		:hasAction                 & :Action                        & 1..*                 & Yes   \\ \hline
		:hasObservationFeature     & :ObservationFeature            & 1..*                & Yes     \\ \hline
		:hasNumberOfActors         & xsd:integer                   & 1..*                & Yes    \\ \hline
		:hasCommunicationType      & \{asynchronous, synchronous\} & 1                & Yes                    \\ \hline
		:isSequential              & xsd:boolean                   & 1                & Yes                    \\ \hline
	\end{tabular} %
	}
\end{table}

\noindent The concept \emph{State} represents the current state of an MDP within an environment. A state has the properties listed in Table~\ref{tab:state}. For example, each state is associated with an arbitrary number of observations that the agent can make. These observations determine the corresponding state. A state can have different functions, such as being an initial state, a final state that completes the activity, or a target state that represents the goals of an activity. Usually, the end state and the goal state are identical, but not necessarily. Each state has a certain desirability and this desirability can be expressed by rewards represented by positive or negative scalar values. In order to recognise a state, a rule expression is needed that defines which criteria must be fulfilled for the respective state. This rule expression is necessary for the simulation function to infer which state is currently prevailing. It is important to note that the rule expression is either provided by a subject matter expert or derived from the activity dataset that serves as the basis for the knowledge graph. Eq.~\ref{eq:rule} shows an example of such a rule expression. The premises of the rule expression are concatenated terms, i.e. in this case observation features, such as \emph{SystolicBloodPressure} and \emph{DiastolicBloodPressure}, which cover certain value ranges and thresholds via comparison operators. If the premise of the expression is true, the inferred state follows, i.e. in the example case \emph{HighBloodPressure}. \\ 

\begin{equation}
\mathsf{SystolicBloodPressure} \geq \mathsf{140} \sqcap \mathsf{DiastolicBloodPressure} \geq \mathsf{80} \rightarrow \mathsf{HighBloodpressure}
\label{eq:rule}
\end{equation}

\begin{table}[!htb]
\centering
\caption{The \emph{State} concept of the proposed ontology.}
\label{tab:state}
\resizebox{\textwidth}{!}{%
\begin{tabular}{|cccc|}
\hline
\multicolumn{4}{|c|}{\textbf{Concept:State}}                                                                                                                    \\ \hline
\multicolumn{1}{|c|}{\textbf{Properties}}    & \multicolumn{1}{c|}{\textbf{Range}}     & \multicolumn{1}{c|}{\textbf{Cardinality}} & \textbf{Mandatory} \\ \hline
\multicolumn{1}{|c|}{:hasObservationFeature} & \multicolumn{1}{c|}{:ObservationFeature} & \multicolumn{1}{c|}{1..*}                 & Yes                \\ \hline
\multicolumn{1}{|c|}{:isInitialState}        & \multicolumn{1}{c|}{xsd:boolean}        & \multicolumn{1}{c|}{1}                    & Yes                \\ \hline
\multicolumn{1}{|c|}{:isFinalState}          & \multicolumn{1}{c|}{xsd:boolean}        & \multicolumn{1}{c|}{1}                    & Yes                \\ \hline
\multicolumn{1}{|c|}{:isGoal}           & \multicolumn{1}{c|}{xsd:boolean}        & \multicolumn{1}{c|}{1}                    & Yes                \\ \hline
\multicolumn{1}{|c|}{:hasReward}             & \multicolumn{1}{c|}{xsd:double}         & \multicolumn{1}{c|}{1}                    & Yes                \\ \hline
\multicolumn{1}{|c|}{:hasExpression}         & \multicolumn{1}{c|}{xsd:string}         & \multicolumn{1}{c|}{1}                    & Yes                \\ \hline
\end{tabular}%
}
\end{table}

\noindent The \emph{ObservationFeature} concept is the most expressive concept within the ontology that contains a lot of information required for the simulation function. Earlier, it was mentioned that observation features constitute each state and therefore, represent sensor measurements in data. Since data is usually described by statistical measures and probability distributions, the statistical characteristics of the data that serves as basis for the simulation function, have to be encapsulated within the \emph{ObservationFeature} concept. \\

\noindent It is presumed in this approach that each observation feature has a value range in which it can occur. This value range has a start and an end point. Moreover, it has to be defined what kind of feature, i.e. numeric, nominal or ordinal, is prevalent, since the feature type determines e.g. whether a preprocessing of data is required and how to interpret the data. The unit is optional and allows transformations, e.g. from Celsius degree to Fahrenheit, between different units, if required. Underlying to every data, a probability distribution is inherently given. Depending on the feature type (numeric, nominal), the prevalence of feature values and the probability function are determined, since the simulation function has to simulate stochastic as well as deterministic environments and therefore requires this information. The \emph{ObservationFeature} concept supports commonly used probability density functions (PDFs) for continuous features and probability mass functions (PMFs)  for discrete features, e.g. Gaussian, Poisson, Binomial, Uniform. Depending on the probability distribution functions, specific statistical parameters, e.g. mean, standard deviation, lambda, success- and failure rate are required in order to compute the probability of any upcoming feature value. To further characterise the observation features that base on the underlying datasets, additional statistical information about the data, e.g. number of experiments, median and mode value are provided in the corresponding concept. \\

\begin{table}[!htb]
\centering
\caption{The \emph{ObservationFeature} concept of the ontology.}
\label{tab:observationFeature}
\resizebox{\textwidth}{!}{%
\begin{tabular}{|cccc|}
\hline
\multicolumn{4}{|c|}{\textbf{Concept:ObservationFeature}}                                                                                                                                                                                    \\ \hline
\multicolumn{1}{|c|}{\textbf{Properties}}         & \multicolumn{1}{c|}{\textbf{Range}}                                                                                     & \multicolumn{1}{c|}{\textbf{Cardinality}} & \textbf{Mandatory} \\ \hline
\multicolumn{1}{|c|}{:hasRangeStart}              & \multicolumn{1}{c|}{xsd:double}                                                                                         & \multicolumn{1}{c|}{1}                   & Yes                \\ \hline
\multicolumn{1}{|c|}{:hasRangeEnd}               & \multicolumn{1}{c|}{xsd:double}                                                                                         & \multicolumn{1}{c|}{1}                    & Yes                \\ \hline
\multicolumn{1}{|c|}{:hasFeatureType}             & \multicolumn{1}{c|}{\{NOMINAL, NUMERICAL, ORDINAL\}}                                                                    & \multicolumn{1}{c|}{1}                    & Yes                \\ \hline
\multicolumn{1}{|c|}{:hasUnit}                    & \multicolumn{1}{c|}{xsd:string}                                                                                         & \multicolumn{1}{c|}{1}                    & No                 \\ \hline
\multicolumn{1}{|c|}{:hasProbabilityDistribution} & \multicolumn{1}{c|}{\begin{tabular}[c]{@{}c@{}}\{NONE, GAUSSIAN, EXPONENTIAL, \\ BINOMIAL, POISSON, UNIFORM\}\end{tabular}} & \multicolumn{1}{c|}{1}                    & No                 \\ \hline
\multicolumn{1}{|c|}{:hasLambda}                  & \multicolumn{1}{c|}{xsd:double}                                                                                         & \multicolumn{1}{c|}{1}                    & No                 \\ \hline
\multicolumn{1}{|c|}{:hasMeanValue}               & \multicolumn{1}{c|}{xsd:double}                                                                                         & \multicolumn{1}{c|}{1}                    & No                 \\ \hline
\multicolumn{1}{|c|}{:hasStandardDeviation}       & \multicolumn{1}{c|}{xsd:double}                                                                                         & \multicolumn{1}{c|}{1}                    & No                 \\ \hline
\multicolumn{1}{|c|}{:hasVariance}             & \multicolumn{1}{c|}{xsd:double}                                                                                         & \multicolumn{1}{c|}{1}                    & No                 \\ \hline
\multicolumn{1}{|c|}{:hasMedian}             & \multicolumn{1}{c|}{xsd:double}                                                                                         & \multicolumn{1}{c|}{1}                    & No                 \\ \hline
\multicolumn{1}{|c|}{:hasModeValue}             & \multicolumn{1}{c|}{xsd:double}                                                                                         & \multicolumn{1}{c|}{1}                    & No                 \\ \hline
\multicolumn{1}{|c|}{:hasNumberExperiments}             & \multicolumn{1}{c|}{xsd:integer}                                                                                         & \multicolumn{1}{c|}{1}                    & No                 \\ \hline
\multicolumn{1}{|c|}{:hasNumberSuccesses}             & \multicolumn{1}{c|}{xsd:integer}                                                                                         & \multicolumn{1}{c|}{1}                    & No                 \\ \hline
\multicolumn{1}{|c|}{:hasSuccessRate}             & \multicolumn{1}{c|}{xsd:double}                                                                                         & \multicolumn{1}{c|}{1}                    & No                 \\ \hline
\multicolumn{1}{|c|}{:hasFailureRate}             & \multicolumn{1}{c|}{xsd:double}                                                                                         & \multicolumn{1}{c|}{1}                    & No                 \\ \hline
\end{tabular}%
}
\end{table}

\noindent The \emph{Transition} concept represents the transition from one state to the next state. If the environment is stochastic, these transitions happen with a certain transition probability depending on the action performed, which can be specified in the \emph{Transition} concept. Depending on the current state and the action performed by the agent, the simulation function can use this dataset-specific information to simulate the same probabilities of state transitions as are implicitly encoded in the data. \\

\begin{table}[!htb]
\centering
\caption{The \emph{Transition} concept of the ontology.}
\label{tab:transition}
\resizebox{\textwidth}{!}{%
\begin{tabular}{|cccc|}
\hline
\multicolumn{4}{|c|}{\textbf{Concept:Transition}}                                                                                                      \\ \hline
\multicolumn{1}{|c|}{\textbf{Property}}         & \multicolumn{1}{c|}{\textbf{Range}} & \multicolumn{1}{c|}{\textbf{Cardinality}} & \textbf{Mandatory} \\ \hline
\multicolumn{1}{|c|}{:hasPreviousState}         & \multicolumn{1}{c|}{:State}         & \multicolumn{1}{c|}{1}                    & Yes                \\ \hline
\multicolumn{1}{|c|}{:hasNextState}             & \multicolumn{1}{c|}{:State}         & \multicolumn{1}{c|}{1}                    & Yes                \\ \hline
\multicolumn{1}{|c|}{:hasAction}             & \multicolumn{1}{c|}{:Action}         & \multicolumn{1}{c|}{1}                    & Yes                \\ \hline
\multicolumn{1}{|c|}{:hasTransitionProbability} & \multicolumn{1}{c|}{xsd:double}     & \multicolumn{1}{c|}{1}                    & Yes                \\ \hline
\end{tabular}%
}
\end{table}

\noindent The concept \emph{Action} represents actions that can be performed by an agent within an activity. Characteristic of this concept is that it can cause an arbitrary number of transitions leading from one state to another. This is because each action is assumed to have an effect on the environment that causes these state transitions. An action can optionally have a certain duration or frequency in which it occurs. \\

\begin{table}[!htb]
\caption{The \emph{Action} concept of the ontology.}
\label{tab:action}
\centering
\resizebox{\textwidth}{!}{%
\begin{tabular}{|cccc|}
\hline
\multicolumn{4}{|c|}{\textbf{Concept:Action}}                                                                                                    \\ \hline
\multicolumn{1}{|c|}{\textbf{Properties}} & \multicolumn{1}{c|}{\textbf{Range}} & \multicolumn{1}{c|}{\textbf{Cardinality}} & \textbf{Mandatory} \\ \hline
\multicolumn{1}{|c|}{:hasEffect}          & \multicolumn{1}{c|}{:Effect}        & \multicolumn{1}{c|}{1..*}                 & Yes                \\ \hline
\multicolumn{1}{|c|}{:hasTransition}          & \multicolumn{1}{c|}{:Transition}        & \multicolumn{1}{c|}{0..*}                 & No                \\ \hline
\multicolumn{1}{|c|}{:hasDuration}        & \multicolumn{1}{c|}{xsd:double}     & \multicolumn{1}{c|}{0..1}                    & No                 \\ \hline
\multicolumn{1}{|c|}{:hasFrequency}       & \multicolumn{1}{c|}{xsd:integer}    & \multicolumn{1}{c|}{0..1}                    & No                 \\ \hline
\end{tabular}%
}
\end{table}

\noindent The concept \emph{Effect} represents effects that have an impact on various observation features in the environment. This concept is only required when knowledge graphs for activities are created manually by a subject matter expert who can define how an action affects environmental states. However, when activity entities are created from datasets, the \emph{Transition} entity provides the information needed by the simulation function to simulate state changes. \\ 

\noindent Effects can be distinguished according to the effect type. For example, an effect can \emph{increase, decrease} feature values, switch binary states \emph{on} or \emph{off}, or \emph{convert} binary states to their opposite. Furthermore, equations can be defined that allow the \emph{computation} of the changes caused by the effect. \\

\begin{table}[!htb]
\caption{The \emph{Effect} concept of the ontology.}
\label{tab:effect}
\centering
\resizebox{\textwidth}{!}{%
\begin{tabular}{|cccc|}
\hline
\multicolumn{4}{|c|}{\textbf{Concept:Effect}}                                                                                                                                                                                               \\ \hline
\multicolumn{1}{|c|}{\textbf{Properties}}    & \multicolumn{1}{c|}{\textbf{Range}}                                                                                         & \multicolumn{1}{c|}{\textbf{Cardinality}} & \textbf{Mandatory} \\ \hline
\multicolumn{1}{|c|}{:hasObservationFeature} & \multicolumn{1}{c|}{:ObservationFeature}                                                                                    & \multicolumn{1}{c|}{1..*}                 & Yes                \\ \hline
\multicolumn{1}{|c|}{:hasImpactType}         & \multicolumn{1}{c|}{\begin{tabular}[c]{@{}c@{}}\{INCREASE, DECREASE, CONVERT, \\ ON, OFF, CONSTANT, COMPUTE\}\end{tabular}} & \multicolumn{1}{c|}{1}                    & Yes                \\ \hline
\multicolumn{1}{|c|}{:hasEquation}           & \multicolumn{1}{c|}{:Equation}                                                                                              & \multicolumn{1}{c|}{0..1}                 & No                 \\ \hline
\end{tabular}%
}
\end{table}

\noindent The concept \emph{Equation} is needed for the calculation of effects on observation features. Thus, any function or equation can be defined as an expression that is parsed by the simulation function to compute the corresponding effect on the respective observation feature. It is worth mentioning that the changeable observation features are the variables in these equations, while parameters are also required that influence the course of the respective function graph. Mainly, these equation expressions are needed for differential equations to represent the change of affected features over time. \\

\begin{table}[!htb]
\caption{The \emph{Equation} concept of the ontology.}
\label{tab:equation}
\centering
\resizebox{\textwidth}{!}{%
\begin{tabular}{|cccc|}
\hline
\multicolumn{4}{|c|}{\textbf{Concept:Equation}}                                                                                                  \\ \hline
\multicolumn{1}{|c|}{\textbf{Properties}} & \multicolumn{1}{c|}{\textbf{Range}} & \multicolumn{1}{c|}{\textbf{Cardinality}} & \textbf{Mandatory} \\ \hline
\multicolumn{1}{|c|}{:hasExpression}      & \multicolumn{1}{c|}{xsd:string}     & \multicolumn{1}{c|}{1}                    & Yes                \\ \hline
\multicolumn{1}{|c|}{:hasParameter}       & \multicolumn{1}{c|}{:Parameter}     & \multicolumn{1}{c|}{1..*}                 & Yes                \\ \hline
\end{tabular}%
}
\end{table}

\noindent As explained earlier, the \emph{parameter} concept is necessary for the \emph{equation} concept because it represents the parameters of an equation. A \emph{Parameter} concept is represented by a name and a numerical value. \\

\begin{table}[!htb]
\caption{The \emph{Parameter} concept of the ontology.}
\label{tab:parameter}
\centering
\resizebox{\textwidth}{!}{%
\begin{tabular}{|cccc|}
\hline
\multicolumn{4}{|c|}{\textbf{Concept:Parameter}}                                                                                                 \\ \hline
\multicolumn{1}{|c|}{\textbf{Properties}} & \multicolumn{1}{c|}{\textbf{Range}} & \multicolumn{1}{c|}{\textbf{Cardinality}} & \textbf{Mandatory} \\ \hline
\multicolumn{1}{|c|}{:hasName}            & \multicolumn{1}{c|}{xsd:string}     & \multicolumn{1}{c|}{1}                    & Yes                \\ \hline
\multicolumn{1}{|c|}{:hasValue}           & \multicolumn{1}{c|}{xsd:double}     & \multicolumn{1}{c|}{1}                    & Yes                \\ \hline
\end{tabular}%
}
\end{table}

\subsection{Deriving Activities as MDPs from datasets}
\label{subsubsec:MDP_Creation}
\noindent To enable the simulation of contexts, i.e. states, and the prediction of action effects in the platform, semantic activity entities have to be either provided by domain experts or derived from datasets that emulate activities of humans or agents. Activities can be described in terms of the probability distribution of states, actions performed, and observations made by agents within a state. The implicitly given probability distribution of attributes (i.e. states, actions, observations) in datasets enables the derivation of MDP models. In the following, it is explained how the platform derives MDPs from operational datasets in order to be able to use these MDPs for the creation of knowledge graphs\footnote{The creation of corresponding knowledge graphs is determined by the underlying ontology concepts and their properties.} and the  simulation of action effects and state changes. \\

\noindent The prerequisite for this approach to work is that agents collect and provide data during their run-time that they perceive from their environment. The basic assumption made for the platform is that agents cooperate by providing their knowledge or experience in the form of (time-series) data. Each row of the dataset has to consist of the following attributes: $S_n, A_n, O_n, R_n, S_{n+1}$, where $S_n$ is an observed state, $A_n$ is an executed action in state $S_n$, $O_n$ is one of the observed features that constitute state $S_n$, $R_n$ is the reward or feedback value received for the executed action $A_n$ in state $S_n$ and $S_{n+1}$ is the subsequent state observed. \\

\begin{figure}[!htbp]
\centering
\includegraphics[width=0.75\textwidth]{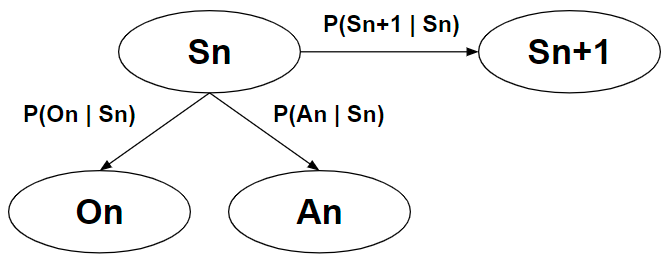}
\caption{A (hidden) Markov model that represents hidden states and state observations.}
\label{fig:hmm}
\end{figure}

\noindent Based on such a dataset and the \emph{Bayes theorem}, transitions, i.e. conditional probabilities, between states and observation features and actions performed can be determined. The goal is to derive a \emph{Hidden Markov Model (HMM)} (see~\cite{Stamp:2004}) that represents the stochastic dynamics of the underlying activity. By deriving Markov chains, the simulation function is able to determine probable state transitions based on observations made and an action performed to simulate the corresponding activity. Thus, depending on a given state $S_n$ and an action performed, the next most likely state can be inferred. Fig.~\ref{fig:hmm} shows which transition probabilities within the HMM are determined and considered by the simulation function. Starting from a state $S_n$, probable transitions to subsequent states are determined by the conditional probability $P(S_{n+1} \mid S_n)$. The conditional probabilities for observations and actions are determined accordingly with $P(O_n \mid S_n)$ for observations and $P(A_n \mid S_n)$ for actions. Based on such a HMM, the expected maximum likelihood of transitions can be estimated either by the \emph{Viterbi}~\cite{Forney:1973} or \emph{Baum-Welch}~\cite{Rabiner:1986} algorithm, which allows the simulation function to simulate the sequence of activities, i.e. the most likely sequences of actions and states for any given observations within the HMM. \\

\noindent For determining the conditional probability and statistical measures of observation features, a distinction has to be made between different types of features. For example, in the case of categorical and binary features, the frequency of occurrence, i.e. the PMF, is counted, whereas in the case of continuous features, a PDF, e.g. the \emph{Gaussian function} (see \cite{Weisstein:2021}), has to be estimated by means of a \emph{kernel density estimation (KDE)}. In these two ways, the conditional probability of each feature in each related state can be determined. Additional statistical measures, e.g. mean, variance, standard deviation, as determined in the MDP ontology can be computed from the corresponding data samples. \\

\noindent Taking into account the obtained reward values that serve as the response for each \emph{state-action-state} transition, the \emph{median} or \emph{mean} reward value obtained for each state is calculated to determine how desirable a state is. \\

\noindent Based on the observations obtained and the current action performed, the simulation function can then determine the most likely subsequent state and corresponding mean reward value. Once the conditional transition probabilities have been determined, an activity instance is created by the platform based on the MDP ontology. Therefore, all instances outlined in Sec.~\ref{sec:ontology} are created by the platform, whose features are obtained from the dataset and correspond to the probability distribution in the HMM. \\

\noindent The created HMM that consists of observed states and performed actions, serves on the one hand as input for knowledge graph generation and on the other hand, as input for the \emph{context-dependent policy composition} approach, so that entity embedding vectors can be trained, as explained in Sec.~\ref{subsubsec:WEV_Pol}. The simulation function uses the resulting activity knowledge graph to provide feedback to the ensemble of agents for their performed actions, see Sec.~\ref{subsec:policies_composition_algorithm}.

\subsection{Training of MDP Entity Embeddings}
\label{subsubsec:WEV_Pol}
\noindent The training of MDP entity embeddings is necessary because different activity contexts have to be numerically encoded to compute their spatial distribution and distance in order to find semantically related properties within a vector space of arbitrary dimension size. The numerical representation enables the computation of semantically related states and actions that occur within an activity. The advantage is that contexts, i.e. the search space of states and actions, can be narrowed down so that semantically related and similarly relevant actions for a current state can be found more quickly. In this way, it is possible to create policies that can lead to the successful execution and completion of an activity. In addition, alternative sequences of actions can be found so that the agent can be offered a variety of ways to perform an activity, depending on the prevailing context.\\

\begin{figure}[!htbp]
\centering
\includegraphics[width=\textwidth]{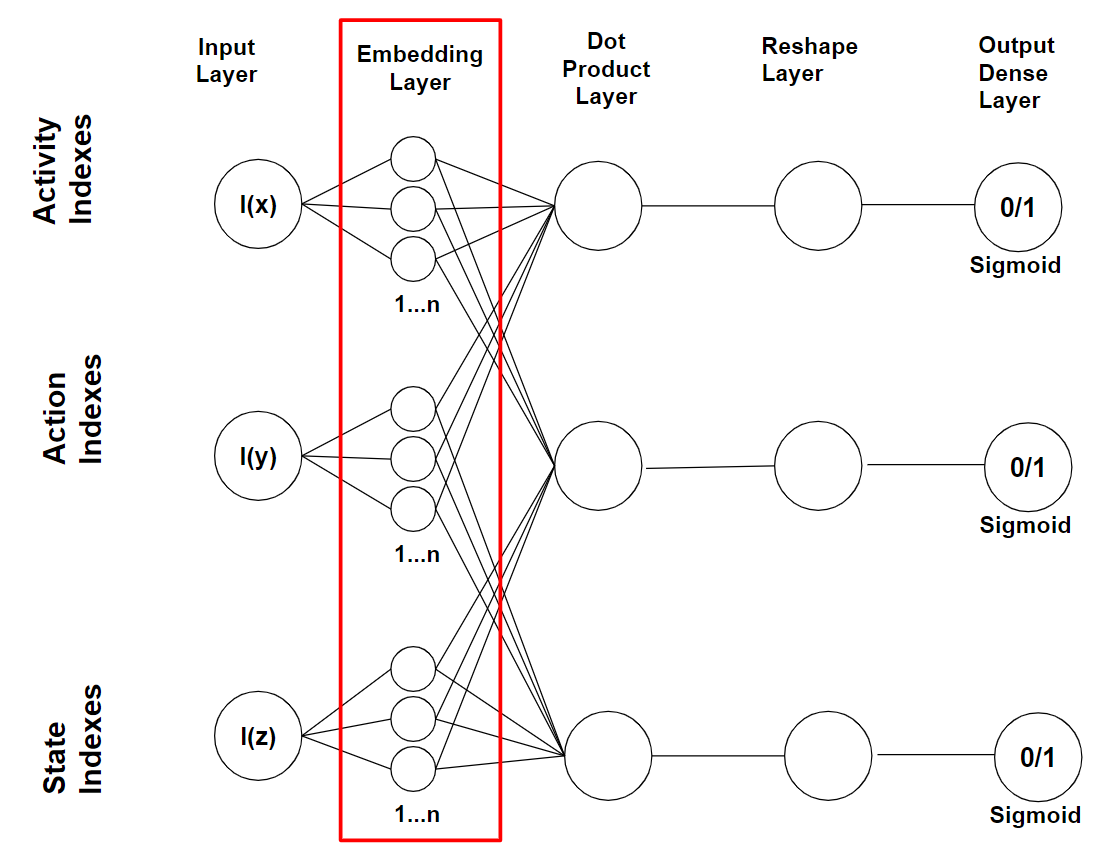}
\caption{Arrangement of the deep neural network for training entity embedding vectors (see layers in red box).}
\label{fig:embeddingNetwork}
\end{figure}

\noindent The data samples that serve as input to the deep neural network (DNN) consist of the following classes: (\emph{activity id, action id, subsequent state id}, see input layer in Fig.~\ref{fig:embeddingNetwork}). The mentioned classes represent binary classes and can therefore either have the value 0 for \emph{no co-occurrence} and 1 for \emph{co-occurrence}. For instance, a target output of a data sample consisting of (Activity$_x$ = 1, Action$_y$$_t$ = 1, State$_z$$_{t+1}$ = 1) indicates that in the sample the referenced action and state occur together within in the given activity. A co-occurrence is determined by the fact that activity entities reference (i.e. link) the corresponding action and state entities, while a co-occurrence between state and action entities exists when an action at time $t$ has directly led to the corresponding subsequent state at time $t+1$. The data samples thus come from two different sources of information, on the one hand the activities and their relationships to the performed actions and on the other hand the consequential states caused by the actions performed. To obtain representative entity embeddings and avoid imbalanced datasets, it is necessary to provide as many positive as negative examples of co-occurring entities. \\

\noindent The concerned entities are encoded by index numbers since a DNN can only process numerical inputs. Therefore, a dictionary for each entity concept has to be maintained that maps the entity names to the corresponding index numbers. In each training iteration, batches of training samples are assembled in order to serve as input for the DNN. The DNN that trains the entity embeddings for each concept, i.e. activity/action, activity/state and action/state, is depicted in Fig.~\ref{fig:embeddingNetwork} and consists of the following network layers: input, embedding, dot product that joins the embedding layer outcomes, a layer for reshaping the dot product outcomes to a shape of one and a fully connected dense layer as output layer that projects by the Sigmoid activation function, the reshaped vector value to an output value between 0 and 1. \\

\begin{figure}[!htbp]
\centering
\includegraphics[width=\textwidth]{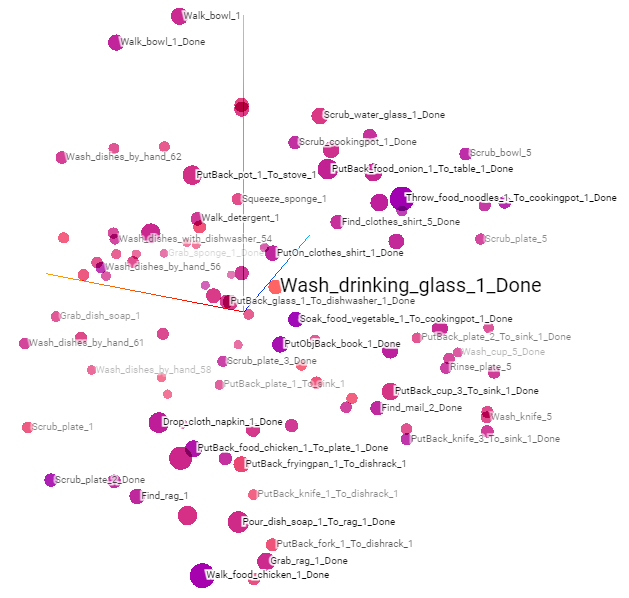}
\caption{Trained embeddings of activities, states and actions that were merged and visualised in a 3-dimensional space in the Tensorflow Projector. The colouring of the embedding vectors (visualised as dots) shows how distant other embedding vectors are from a selected embedding vector (here, e.g. in light red \emph{Wash\_drinking\_glass\_1\_Done}). The darker the hue of a neighbouring embedding vector, the closer it is to the selected vector. The distances in this figure were measured with the Euclidean distance.}
\label{fig:VisualEmbeddings}
\end{figure}

\noindent The embedding layer can consist of any number of units resp. neurons, but the evaluation results in this work indicate that 50 units resp. neurons are sufficient to achieve convincing results. Optimisation of the embedding layers was performed using the \emph{Adam optimiser}~\cite{Kingma:2015} algorithm was used. However, other optimisation algorithms (e.g. stochastic gradient descent) are also possible. It is important to note that for each entity of activities, states and actions, an embedding vector is trained that represents the corresponding entity in the n-dimensional embedding vector space. Thus, the main goal is to obtain representative embedding vectors that indicate the semantic relatedness of entities through their observed context. Thus, the main result of the DNN presented are the numerical word embedding vectors of an arbitrary dimension (here 50) representing each MDP entity in an n-dimensional vector space (see exemplary embedding vector below) and an excerpt of its visual representation in Fig.~\ref{fig:VisualEmbeddings}. \\

$
\vec(embedding_n)=\begin{pmatrix}
	0.344394855487582 \\
	-0.35454000765544 \\
	0.339430778676755 \\
	... \\
	n \\
\end{pmatrix}
$
\\

\noindent As loss function that measures the error rate during training, we utilised \emph{cross-entropy} since it is intended for binary classification problems as required in the proposed DNN. The obtained result of the DNN are the trained weights of the embedding layer, i.e. the embedding vectors for each entity of the given dataset. These embedding vectors are consolidated and stored in a TSV file, while their indices and human-readable names are stored in a separate TSV file. \\

\noindent One could argue that conditional probabilities could also be used to obtain the most likely strategies for a given state. However, compared to conditional probabilities, the advantage of entity embeddings is that they allow semantic properties, i.e. relationships between different entities, to be revealed and, as mentioned earlier, arithmetic operations to be performed that reveal additional features of entity relationships~\cite{Allen:2019}. For example, it is possible to add or subtract embedding vectors, which can lead to new, closely related entity vectors that have a certain semantic meaning. In contrast to embeddings, conditional probabilities do not capture the semantic relationships between different entities. 

\subsection{The Simulation Function}
\label{sec:simulation}
\noindent Each ensemble agent implements and invokes within its program thread a simulation function that receives from the agent the initial parameters. It is important to note that the simulation function is a \emph{closure function}, which is a known concept from functional programming. When the simulation function is called for the first time, the SPARQL endpoint reference URL to the activity knowledge graph as well as the initial state are passed to it, so that the function can access the knowledge graph and knows how to handle state updates and feedback to the agent \textbf{(line 1)} of Algo~\ref{algo:SimulationFunc}. The fact that the simulation function is a \emph{closure function} that returns an inner function \textbf{(line 4)}, allows the agent to invoke the function after its initialisation \textbf{(lines 2-3)}, repeatedly in each single simulation step. The simulation function manages and updates the current state using the following steps, see Algo.~\ref{algo:SimulationFunc}. It parses the rule expression of the current state to infer the corresponding state label \textbf{(line 5)}. With the obtained state label, it searches the knowledge graph for the most likely transition entity that links the current state and the action performed \textbf{(line 6)}. To do this, it applies a \emph{SPARQL CONSTRUCT} query that constructs all requested information, i.e. statements, from the knowledge graph. Depending on the transition probability, and the statistical information provided in the observation feature entities, the simulation function updates the current \textbf{(line 8)} state and returns the new updated state together with its referenced reward value to the agent \textbf{(line 9)}. These steps are repeated until the simulation function sends a final state to the corresponding agent instance, which then terminates the simulation. As soon as all running agents have terminated their simulation function, the process ends and the \emph{Agent Ensemble Generator} sends back to the requesting agent the assembled policies ranked by their performance (see Table~\ref{tab:policies}), i.e. obtained reward values. 

\begin{algorithm}
\SetKwInput{KwInput}{Input}                
\SetKwInput{KwOutput}{Output}              
\DontPrintSemicolon
  \KwInput{initialState, sparqlEndpoint}
  \KwOutput{InnerFunction}
  \SetKwFunction{FSimulation}{SimulationFunction}
 
  \SetKwProg{Fn}{Function}{:}{}
  \Fn{\FSimulation{initialState, sparqlEndpoint}}{
  currentState = initialState\;
  endpoint = sparqlEndpoint \;
  \KwRet \Fn{{(performedAction)}}{
 	stateLabel = parseRuleExpression(currentState) \;
	matchingTransitions = lookUpTransition(endpoint, stateLabel, performedAction) \;
	updatedState = updateCurrentState(matchingTransitions) \;
	currentState = evaluateExpression(updatedState) \; 
  	\KwRet currentState \;
  }\;
}
  \caption{SimulationFunction}
  \label{algo:SimulationFunc}
\end{algorithm}

\subsection{Policies Composition Algorithm}
\label{subsec:policies_composition_algorithm}
\noindent Once the entity embedding vectors are trained, policies can be composed by ensembles of agents based on the sent state information of a requesting agent. The algorithm for composing state-based policies is executed by the \emph{AgentEnsembleGenerator} and shown in Algo.~\ref{algo:PoliciesComposition}. The input parameters required for the algorithm are the current state representation sent by the requesting agent and the trained entity embedding vectors of each state and action entity referenced in the corresponding KG \textbf{(line 1)}. The returned output of the algorithm is an array of policies. The algorithm tracks and stores the current state, which is the starting point for each ensemble agent \textbf{(line 3)}. To store the policies found, a data structure, i.e. an array, is declared \textbf{(line 4)}. Then, a \emph{while-loop} is started, which stops as soon as a goal state is reached that terminates the current activity \textbf{(line 5)}. The \emph{while-loop} starts with the selection of the embedding vector representing the sent state \textbf{(line 6)}. The embedding vector of the state is needed to find the nearest embedding vectors of the nearest actions \textbf{(line 7)}. Therefore, a similarity metric (e.g. cosine distance, Euclidean distance, Manhattan distance) is used to calculate the distance between embedding vectors in an n-dimensional embedding vector space. An array stores, for each agent in the ensemble, its feedback received from the simulation function (i.e. the updated state and the reward received) \textbf{(line 8)}. \\

\noindent For each action that is within the specified search radius, a new agent is created and initialised with the corresponding state and the action to be performed \textbf{(line 8 - 10)}. Then, for each agent an instance of the function \emph{simulate} is called, which executes the respective action of each ensemble agent \textbf{(line 11)}. The function \emph{simulate()} then returns the updated state with the corresponding reward value. Afterwards, the updated state object is stored in the results array that is maintained by the \emph{EnsembleGenerator} \textbf{(line 12)}. This is done within a \emph{for-loop} for each agent of the ensemble \textbf{(line 9-12)}. \\

\noindent The advantage of using agent ensembles is that actions can be executed in parallel threads and each thread with its own simulation function updates and manages its own state. The ensemble agents are then synchronised in the main thread by selecting the best, i.e. reward maximising, state to continue the MDP and initialise the next generation of ensemble agents. \\

\noindent After all ensemble agents have performed their assigned actions, a function called \emph{selectBestResultByReward()} is called that sorts all result states in descending order by their reward values \textbf{(line 14)}. The action that achieved the highest reward value compared to the last reward value is then selected as the best action and the \emph{policies} array stores this action together with the corresponding previous state and updates the variable \emph{currentState} to the new state that provided the best result \textbf{(line 17-19)}. In addition, the variable \emph{radius} is reset to the originally set value of \emph{maxDistance}, since only the closest actions should be considered again for the next policy search, which saves computing time and resources, since the number of actions to be performed by agents increases or decreases proportionally to the search radius of the state \textbf{(line 20)}. However, if the current rewards received are less than or equal to the last best reward, then the variable \emph{radius} is increased by 0.25 each time the algorithm gets stuck and does not return reward-maximising actions \textbf{(line 15-16)}. The parameter named \emph{maxDistance} is thus a hyper-parameter and initially defines the radius boundary within which the nearest action embedding vectors have to be located \textbf{(line 2)}. The function exits and returns the array of policies once the target state of the current activity is reached \textbf{(line 22)}. \\

\begin{algorithm}
\SetKwInput{KwInput}{Input}                
\SetKwInput{KwOutput}{Output}              
\DontPrintSemicolon
  \KwInput{initialState, stateEmbeddings, actionEmbeddings, maxDistance}
  \KwOutput{policies}
  \SetKwFunction{FPoliciesComposition}{PoliciesComposition}
 
  \SetKwProg{Fn}{Function}{:}{}
  \Fn{\FPoliciesComposition{initialState, stateEmbeddings, actionEmbeddings, maxDistance}}{
  radius = maxDistance \;
  currentState = initialState\;
  policies = [] \;
  \While{!currentState.isGoal} {
  stateVector = stateEmbeddings[currentState]\;
  closestActions = findClosestActions(stateVector, actionEmbeddings, radius)\;
  	results = []\;
  	\ForAll{action IN closestActions}{
  	    agent = new Agent()\;
  	    updatedState = simulate(agent.exec(currentState, action))\;
  	    results.push(updatedState)
  	}\;
  	bestResult = selectBestResultByReward(results)\;
  	\If{bestResult.reward <= currentState.reward} {
  	    radius += 0.25 \;
  	}
  	\Else{
  	    policies.push(bestResult.state, bestResult.action)\;
  	    currentState = bestResult\;
  	    radius = maxDistance\;
  	}
  	} \;
  	\KwRet policies \;
  }
  \caption{PoliciesComposition}
  \label{algo:PoliciesComposition}
\end{algorithm}

\subsection{Limitations in the Approach}
\noindent In the approach presented, it is assumed that state labels are provided in the datasets used for knowledge graph generation. This precondition implies that human annotators provide the appropriate state labels to the corresponding actions and observation features in the data, as described in Sec.~\ref{subsubsec:MDP_Creation}. Before the presented approach can work, it is a requirement that each new activity record is automatically transformed into a knowledge graph and encoded into corresponding entity embeddings, so that both are available for all possible contexts. Furthermore, it is assumed that the knowledge graph and entity embeddings database will evolve over time and be enriched with new activity entities and entity embeddings, so that queries can be made by agents as needed depending on the stored context information. This implies that the DNN requires to be trained with the new data, i.e. the new activities, states and actions that are added. The advantage is that this provided knowledge is continuously built up so that heterogeneous contexts can be described semantically and used for simulation and policy search purposes.  

\section{Evaluation}
\label{sec:evaluation}
\noindent The evaluation carried out is intended to prove the hypothesis raised and answer the RQs posed. In this section we first present the experimental set-up and then discuss the results and limitations of the evaluation.

\subsection{Experimental Set-up}
\noindent As mentioned earlier, the descriptions of domestic activities provided by the VH platform serve as the dataset for training the required entity embedding vectors. In addition, the VH dataset represents the \emph{ground truth} against which the success of the composite policies is tested. Listing~\ref{lst:data} shows the general structure of the VH dataset and Listing~\ref{lst:sample} depicts an exemplary activity (\emph{Watch\_TV\_49}) for watching TV. The VH dataset consists of activity names \textbf{(line 1)}, a textual description of the activity \textbf{(line 2)} and sequential actions \textbf{(line 4-7)} that are executed in rooms on domestic objects (e.g. furniture, devices). Since the dataset under consideration does not provide named state entities, we had to construct semantic entity descriptions (see Listing~\ref{lst:turtle}) that also contain (subsequent) states and serve as the basis for the simulation function and the DNN used. For instance, if the action \emph{Walk\_living\_room\_1} prevails, then the subsequent artificially constructed state name would be \emph{Walk\_living\_room\_1\_Done}. The state name indicates that the named action has been performed, i.e. done, and a new state is reached. The semantic entity descriptions are based on the concepts of the MDP ontology mentioned in Sec.~\ref{sec:ontology}. Thus, the example in Listing~\ref{lst:turtle} exemplifies the activity \emph{Watch\_TV\_49} with one: a) referenced state (\emph{Walk\_living\_room\_1\_Done}), b) observation feature (\emph{IsWalk\_living\_room\_1}), c) action,  (\emph{Walk\_living\_room\_1}), d) transition with a (\emph{UUID}) and e) effect (\emph{SetWalk\_living\_room\_1}) each. For space reasons, it was avoided to list all entities linked within the mentioned activity. Some activities of the VH dataset have redundant names, but differ in some of their actions and in the order in which the actions are performed. In order to make all activities of the VH dataset unique and unmistakable, we have extended the activity names with consecutive numbers and added to each activity an initial- and final state that both indicate the beginning and completion of the activity. \\

\begin{lstlisting}[language=SPARQL, numbers=left, label=lst:data, caption={General structure of the Virtual Home dataset}]
	Name of the activity
	Natural language description of the activity
	
	[action 1] <object> (index of object)
	[action 2] <object> (index of object)
	...
	[action n] <object> (index of object)
\end{lstlisting}

\begin{lstlisting}[language=SPARQL, numbers=left, label=lst:sample, caption={Watch\_TV\_49 data sample}]
Watch TV
walk to living room, find couch, sit on couch, find remote control, 
turn on tv by pressing button


[Walk] <living_room> (1)
[Walk] <couch> (1)
[Find] <couch> (1)
[Sit] <couch> (1)
[Find] <remote_control> (1)
[Find] <television> (1)
[TurnTo] <television> (1)
\end{lstlisting}

\begin{lstlisting}[language=SPARQL, numbers=left, label=lst:turtle, caption={The semantic entity description of the activity named Watch\_TV\_49 in Turtle format.}]
	# Prefix, i.e. Namespace, Definitions
	@prefix entity: <http://example.org/Entity/> .
	@prefix property: <http://example.org/Property/> .
	@prefix concept: <http://example.org/Concept/> .
	@prefix xsd: <http://www.w3.org/2001/XMLSchema#> .
	@prefix rdf: <http://www.w3.org/1999/02/22-rdf-syntax-ns#> .
	@prefix rdfs: <http://www.w3.org/2000/01/rdf-schema#> .
	
	# Activity Entity (Watch_TV_49) (Mandatory)
	entity:Watch_TV_49 a concept:Activity;
	property:isSequential "true"^^xsd:boolean;
	property:hasNumberOfActors "1"^^xsd:integer;
	property:hasCommunicationType "Asynchronised"^^xsd:string;
	property:hasState entity:Walk_living_room_1_Done;
	property:hasState entity:Walk_couch_1_Done;
	property:hasState entity:Find_couch_1_Done;
	property:hasState entity:Sit_couch_1_Done;
	property:hasState entity:Find_remote_control_1_Done;
	property:hasState entity:Find_television_1_Done;
	property:hasState entity:TurnTo_television_1_Done;
	property:hasState entity:InitialState_Watch_TV_49;
	property:hasState entity:FinalState_Watch_TV_49;
	property:hasObservationFeature entity:IsWalk_living_room_1;
	property:hasObservationFeature entity:IsWalk_couch_1;
	property:hasObservationFeature entity:IsFind_couch_1;
	property:hasObservationFeature entity:IsSit_couch_1;
	property:hasObservationFeature entity:IsFind_remote_control_1;
	property:hasObservationFeature entity:IsFind_television_1;
	property:hasObservationFeature entity:IsTurnTo_television_1;
	property:hasAction entity:Walk_living_room_1;
	property:hasAction entity:Walk_couch_1;
	property:hasAction entity:Find_couch_1;
	property:hasAction entity:Sit_couch_1;
	property:hasAction entity:Find_remote_control_1;
	property:hasAction entity:Find_television_1;
	property:hasAction entity:TurnTo_television_1;
	
	# An Exemplary State Entity (Mandatory)
	entity:Walk_living_room_1_Done a concept:State;
	property:isGoal "false"^^xsd:boolean;
	property:isFinalState "false"^^xsd:boolean;
	property:isInitialState "false"^^xsd:boolean;
	property:hasExpression "IsWalk_living_room_1 == 1"^^xsd:string;
	property:hasReward "0"^^xsd:double;
	property:hasObservationFeature entity:IsWalk_living_room_1;
	property:hasAction entity:Walk_living_room_1 .
	...
	
	# An Exemplary Observation Feature Entity (Mandatory)
	entity:isWalk_living_room_1 a concept:ObservationFeature;
	property:hasRangeStart "0"^^xsd:double;
	property:hasRangeEnd "1"^^xsd:double;
	property:hasFeatureType "NOMINAL"^^xsd:string;
	property:hasUnit ""^^xsd:string .
	...
	
	# An Exemplary Action Entity (Mandatory)
	entity:Walk_living_room_1 a concept:Action;
	property:HasTransition entity:bec16c1e-b08e-496b-ba10-95e85be65fb9;
	property:HasEffect entity:SetWalk_living_room_1.
	...
	
	# An Exemplary Transition Entity (Optional)
	entity:4cd8f07f-c79f-45f9-b872-25b8cbb0e42f a concept:Transition;
	property:HasPreviousState entity:Walk_living_room_1_Done;
	property:HasNextState entity:Walk_couch_1_Done;
	property:HasAction entity:Walk_couch_1;
	property:HasTransitionProbability "1"^^xsd:double.
	...
	
	# An Exemplary Effect Entity (Mandatory)
	entity:SetWalk_living_room_1 a concept:Effect;
	property:hasImpactType "ON"^^xsd:string;
	property:hasObservationFeature entity:IsWalk_living_room_1 .
	...
\end{lstlisting}

\noindent All elements of the data samples, i.e. activity, action and generated subsequent states, are considered as vocabulary of the dataset and are given a unique index number which serves as input for the DNN. For instance, if the activity named \emph{Watch\_TV\_49} has ID 1 and the action named \emph{Walk\_Livingroom} has ID 2, then the input value for the corresponding DNN is 1 for the activity input and 2 for the action input. During the training process, the corresponding DNN will determine if both IDs occur together in the training records. In this way, the implemented DNN has trained entity embedding vectors for all VH activities, states and actions within 1000 iterations. In each iteration, 15 training epochs were executed and a \emph{generator function} assembled a batch of 1024 positive, i.e. co-occurring, and negative, i.e. non co-occurring, samples for each entity combination.  \\

\begin{figure}[!htbp]
    \centering
    \includegraphics[width=\textwidth]{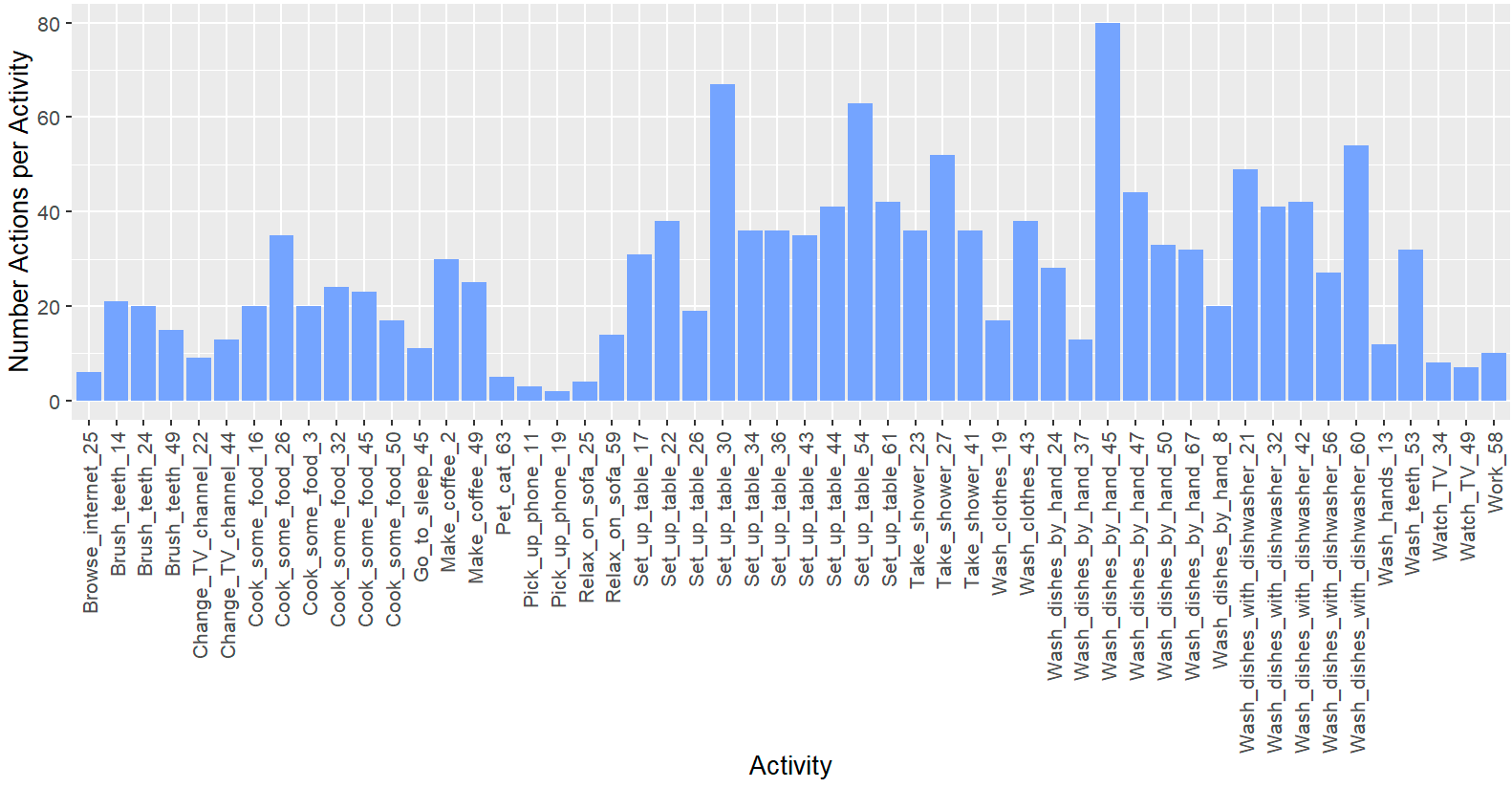}
    \caption{Distribution of the action sequence length (y-axis) among all 52 activities (x-axis) evaluated.}
    \label{fig:num_policies_distibution}
\end{figure}

\noindent In order to obtain representative samples that offer activities of varying complexity, we categorised the data samples (i.e. the activities) according to the length of their action sequences and randomly selected a certain number of samples from each category. The reason for this categorisation is that the complexity and difficulty of an activity increases as the length of the action sequence increases. Since there were a total of 52 different action sequence lengths and thus categories, we decided to randomly select one activity from each sequence length category, resulting in a sample size of 52 activities that were assessed. The action sequence lengths of an activity can range from a minimum of 2 actions to a maximum of 80 actions (see Fig.~\ref{fig:num_policies_distibution}). \\

\noindent To compare RL with our approach in terms of policy delivery speed, we used a JavaScript library\footnote{\url{https://github.com/karpathy/reinforcejs}} implementing the DQNN algorithm~\cite{Mnih:2015} and trained it for each previously selected activity. We decided to use DQNN as the baseline algorithm because it is a representative, i.e., state-of-the-art, algorithm that is widely used, especially for tasks with large state spaces. Alternatively, we could have used, e.g., the \emph{policy gradient}~\cite{Peters:2010} approach, but even with this algorithm, different states and action paths have to be traversed in numerous iterations to obtain strategies that lead to a reward-maximizing outcome. Regardless of the RL algorithm evaluated, the speed and success of RL strongly depend on the complexity, i.e., the size of the action and state space. Moreover, the state space in any RL approach is much larger than in our approach because we constrain the state space by identifying and assigning semantically related actions and states. \\

\noindent Our approach computed policies for the same activities and both approaches monitored the required number of \emph{episodes, execution steps} and the \emph{number of incorrectly executed actions where the reward is < 0} until reward-maximising policies contributing to the completion of an activity were provided. \\

\noindent We initialised the DQNN with the following hyper-parameters: \textbf{learning rate = 0.05}, \textbf{greedy\footnote{Specifies the rate of randomly executed actions and thus controls the exploration and exploitation of the policies.} value = 0.9} and \textbf{discount factor = 0.9}. To determine whether the corresponding policies were successfully learned by the DQNN, the greedy parameter was set to 0 after each training episode to avoid the agent performing random actions, and then the activity was executed again with the trained policies. Then the number of executions performed within an episode and the actual sequence length of the current activity were compared, and if both were equal, the policy learning and training of the DQNN was considered successfully completed. The maximum number of episodes within which the DQNN had to train policies was limited to 1, 10 and 100 episodes, because without episode limitation it would have taken several days to successfully train the policies for all 52 activities with the DQNN. In addition, we want our approach to be able to address contextual policy requests immediately or quickly, and therefore the DQNN used for comparison with our approach also had to meet the requirement of delivering policies quickly. For this reason, the DQNN algorithm was challenged to learn policies for each activity within 1, 10 and 100 episodes. Otherwise, if the number of episodes was exceeded, the training for the corresponding activity was terminated and considered incomplete. \\

\begin{figure}[!htbp]
    \centering
    \includegraphics[width=\textwidth]{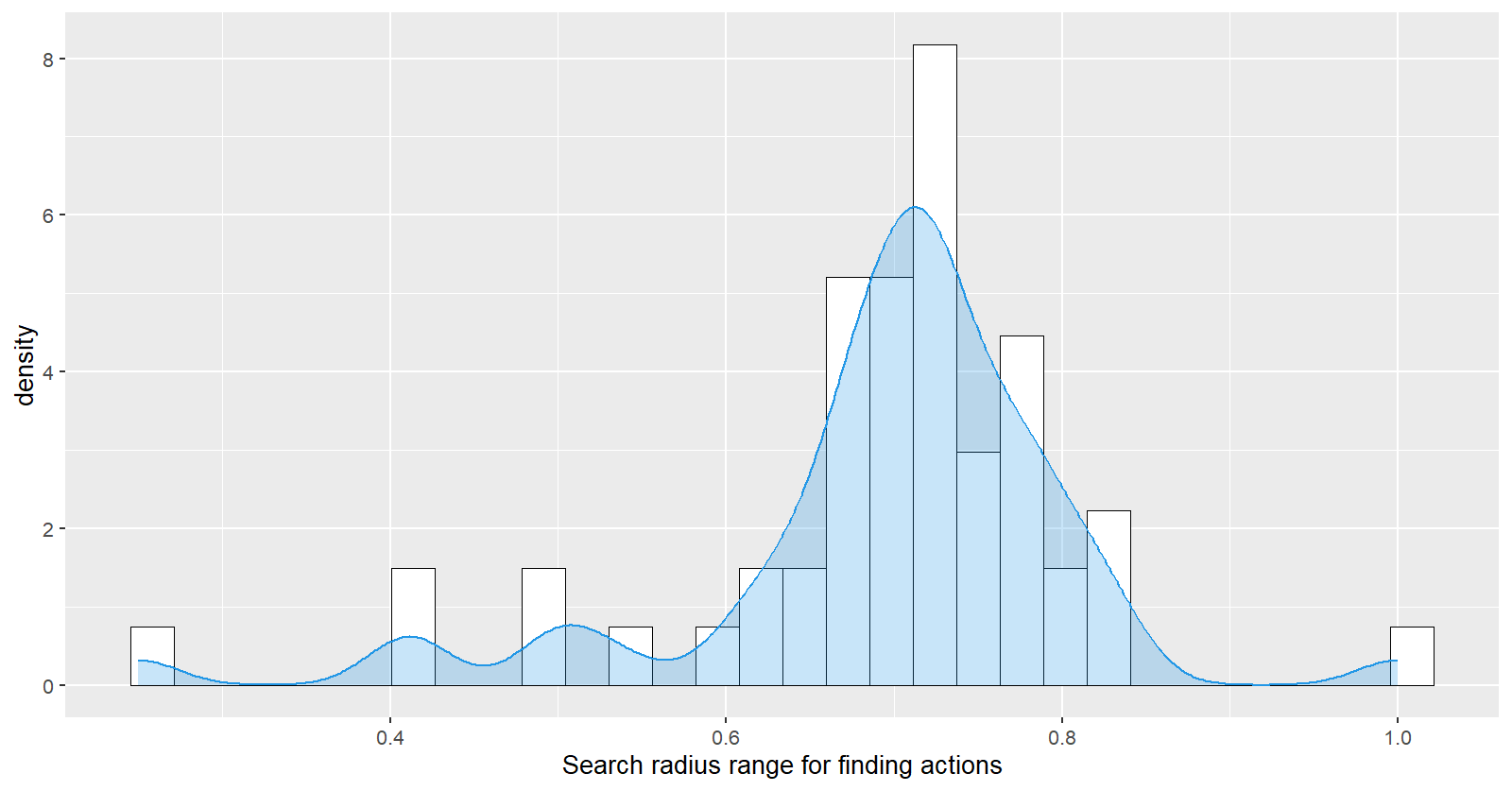}
    \caption{Density plot and histogram of the search radius value ranges (x-axis) in which the most suitable action embeddings for submitted, i.e. observed states, were found.}
    \label{fig:radius}
\end{figure}

\noindent Our proposed algorithm has one hyper-parameter that represents the search radius in the embedding space (see Algorithm~\ref{algo:PoliciesComposition}). During the evaluation, the radius parameter was initially set to \textbf{0.25}, and if no actions were found within this radius, the radius was increased until actions could be found. Fig.~\ref{fig:radius} shows the obtained density plot of radius values over all evaluated activities. It can be seen that the mean search radius value that contained actions matching the current state was in a range between 0.6 and 0.8. Mainly within this search radius, actions that led to reward maximization were found by the algorithm. \\

\noindent The simulation function used for providing feedback (i.e. state-updates, rewards), allocates rewards that either decrease by a value of 0.25 for an incorrectly performed action or increase by a value of 0.25 for each correctly performed action within the action sequence. This incremental reward allocation is necessary for fixed sequences of actions because the algorithm can repeat actions infinitely often and thus get stuck in cycles if states within a sequence are not distinguishable by their increasing and thus different rewards. \\

\subsection{Results}
\noindent The results of the experiments are explained in the following two subsections. The first subsection deals with the experiments and results for RQ 1, the second subsection with the experiments and results for RQ 2. 

\subsubsection{RQ1: Activity Completion}
\begin{figure*}[!htbp]
    \centering
    \begin{subfigure}{.45\textwidth}
    \includegraphics[width=\textwidth]{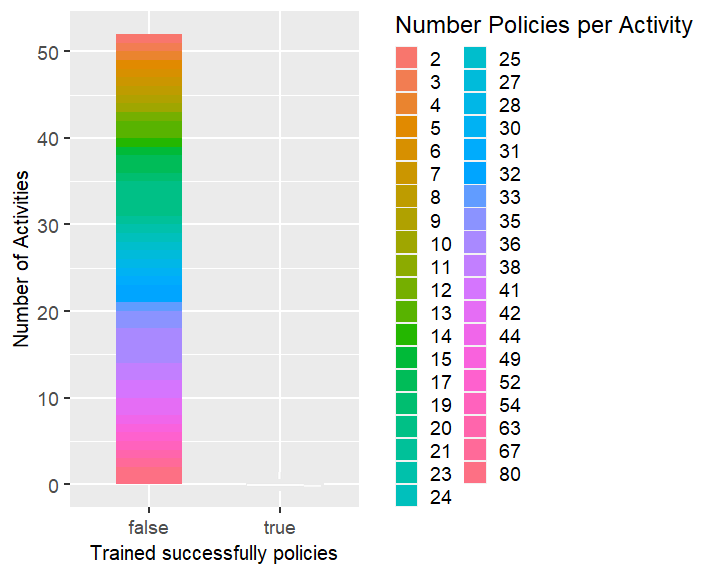}
    \caption{1 Episode}
    \label{fig:hist_episode_1}
    \end{subfigure}
    \begin{subfigure}{.45\textwidth}
    \includegraphics[width=\textwidth]{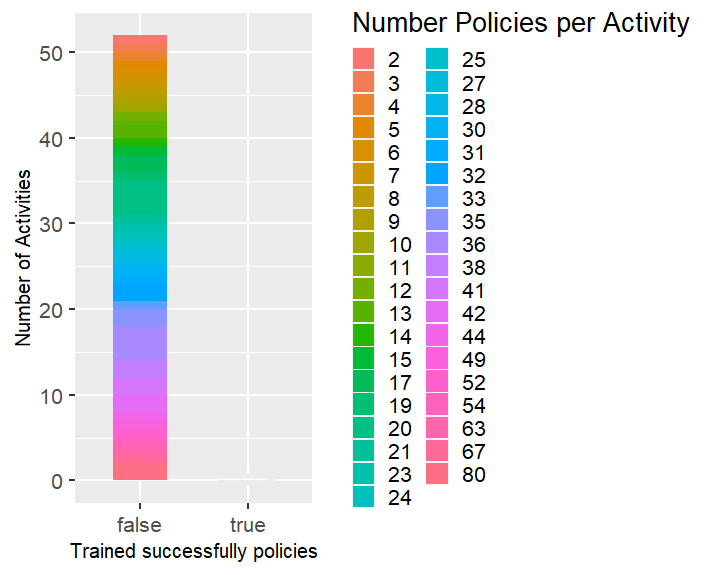}
    \caption{10 Episodes}
    \label{fig:hist_episode_10}
    \end{subfigure}
    \begin{subfigure}{.45\textwidth}
    \includegraphics[width=\textwidth]{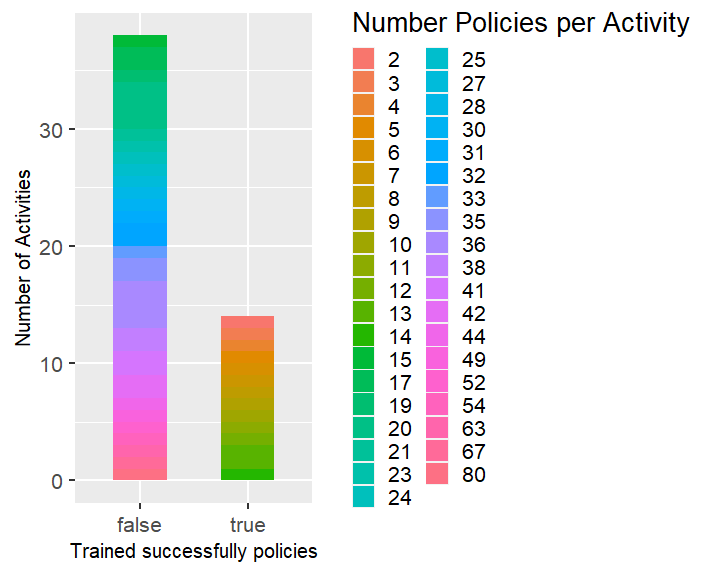}
    \caption{100 Episodes}
    \label{fig:hist_episode_100}
    \end{subfigure}
    \begin{subfigure}{.45\textwidth}
    \includegraphics[width=\textwidth]{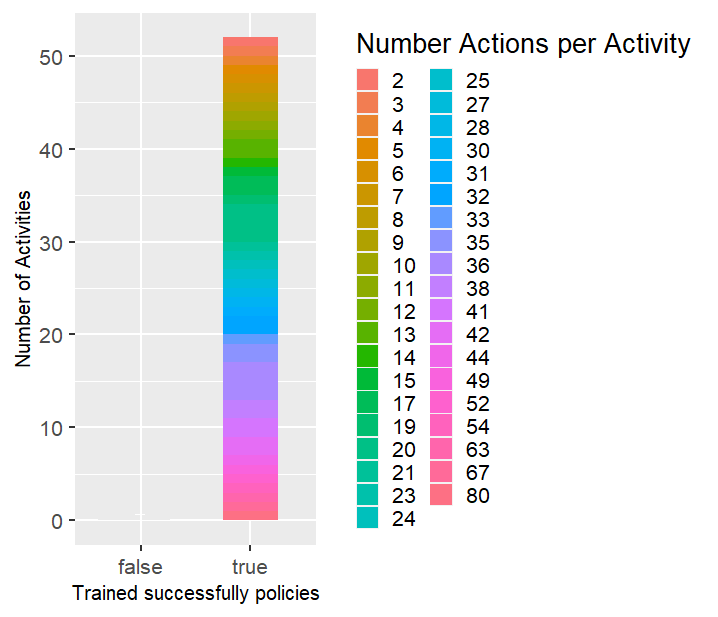}
    \caption{Ensemble Agent approach in 1 Episode}
    \label{fig:hist_ensemble_episode_1}
    \end{subfigure}
    \caption{Distribution of successfully trained (right bar) and unsuccessfully learned (left bar) activities by RL agents after 1, 10 and 100 training episodes and the ensemble agent approach after 1 episode. The colour shades indicate the action sequence length of the respective activities.}
    \label{fig:hists_success}
\end{figure*}

\noindent The evaluation results show that the proposed approach was able to successfully generate policies for all 52 activities within one episode each, while the DQNN agent needed at least 100 episodes to learn policies for 14 out of 52 activities (see Fig.~\ref{fig:hists_success}). For the remaining 38 activities, the DQNN agent would have needed many more episodes before learning reward-maximising and correct sequential behaviour, because as the action sequence length increases, the state space and action choices also increase, making it more difficult and time-consuming for the agent to learn correct action sequences. Fig.~\ref{fig:hists_success} illustrates that the DQNN agent was not able to successfully learn any of the activities within 1 and 10 episodes, while after 100 episodes it had learned at least 14 activities with minimum and maximum action sequence lengths of 2 and 14. For activities with a sequence length greater than 14, the DQNN agent failed to learn the correct action sequence within the 100 episodes. \\

\begin{figure*}[!htbp]
    \centering
    \begin{subfigure}{.33\textwidth}
    \includegraphics[width=\textwidth]{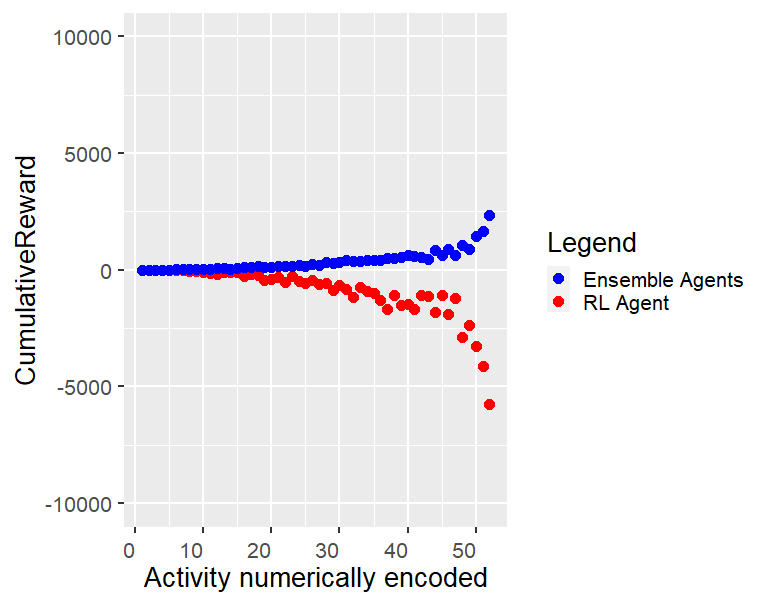}
    \caption{1 Episode}
    \label{fig:cum_reward_1}
    \end{subfigure}
    \begin{subfigure}{.33\textwidth}
        \includegraphics[width=\textwidth]{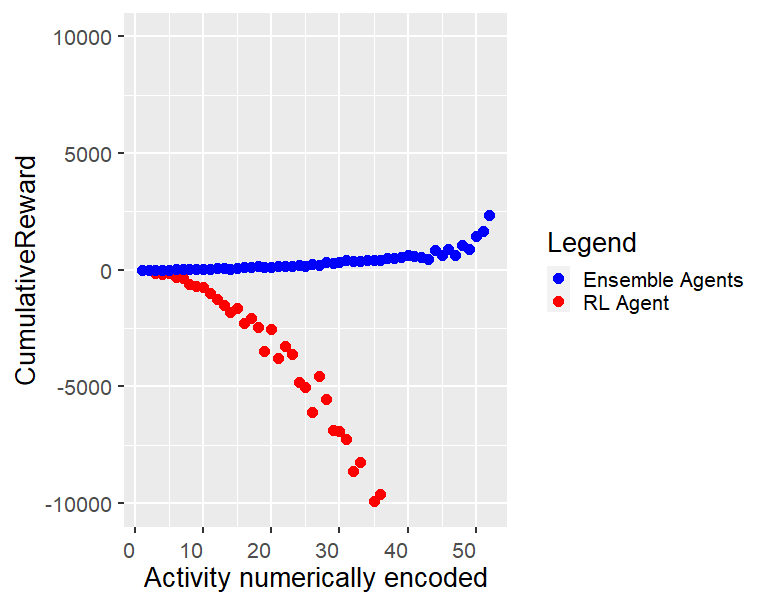}
        \caption{10 Episodes}
        \label{subfig:mixed_scatter_10}
    \end{subfigure}
    \begin{subfigure}{.33\textwidth}
        \includegraphics[width=\textwidth]{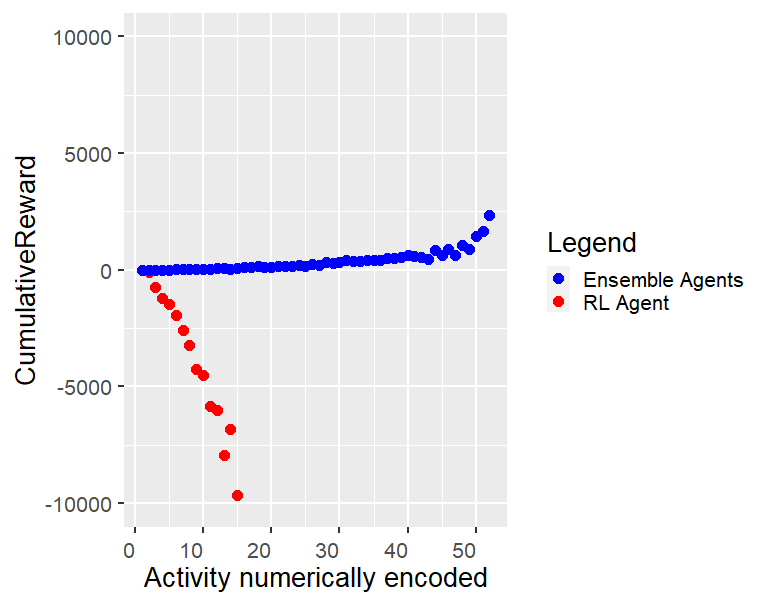}
        \caption{100 Episodes}
        \label{subfig:mixed_scatter_100}
    \end{subfigure}
    \caption{Cumulative reward (y-axis) for each activity (x-axis) obtained during composition by ensemble agents (blue dots) and during training by RL agent (red dots) after 1, 10 and 100 Episodes. It is striking that ensemble agents consistently accumulate positive cumulative reward across all activities assessed. For visualisation reasons, the y-axis had to be limited to a range between -10000 and 10000. Therefore, some data points are not shown in the graphs because they are outside the range shown. However, the trend of the data points remains constant and it becomes obvious that the measured values diverge strongly.}
    \label{fig:plots_cum_rewards}
\end{figure*}

\noindent Fig.~\ref{fig:plots_cum_rewards} shows for our approach (blue dots) and for the DQNN agent (red dots) the cumulative rewards obtained during policy learning and composition. The scatter plots prove that the proposed ensemble agent approach shows reward maximising behaviour and thus composes the right policies, while the DQNN agent tends to show a reward minimising behaviour. This is reasonable because the principle of RL is to learn policies by \emph{trial and error} and therefore the DQNN agent performs a certain number of exploration steps depending on the greedy value set, which can lead to wrong decisions and negative rewards, and since the DQNN agent needs many episodes to learn policies, the penalties add up.  \\

\noindent As described before, 52 different activities with different action sequence lengths and thus difficulty levels were evaluated and for all 52 activities the presented approach was able to compose VH dataset compliant policies. Therefore, it can be concluded for \textbf{RQ1} that our approach is able to assemble reward-maximising policies across contexts, i.e. across activities, that contribute to the fulfilment of activities without having to train policies in advance.

\subsubsection{RQ2: Velocity of Policies Provision}
\begin{figure*}[!htbp]
    \centering
    \begin{subfigure}{.33\textwidth}
            \includegraphics[width=\textwidth]{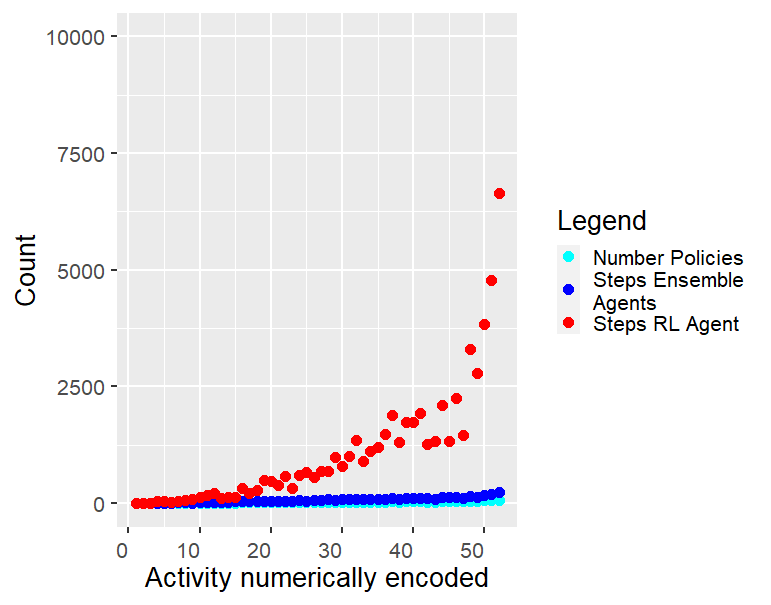}
            \caption{1 Episode}
            \label{fig:mixed_1}
    \end{subfigure}
    \begin{subfigure}{.33\textwidth}
            \includegraphics[width=\textwidth]{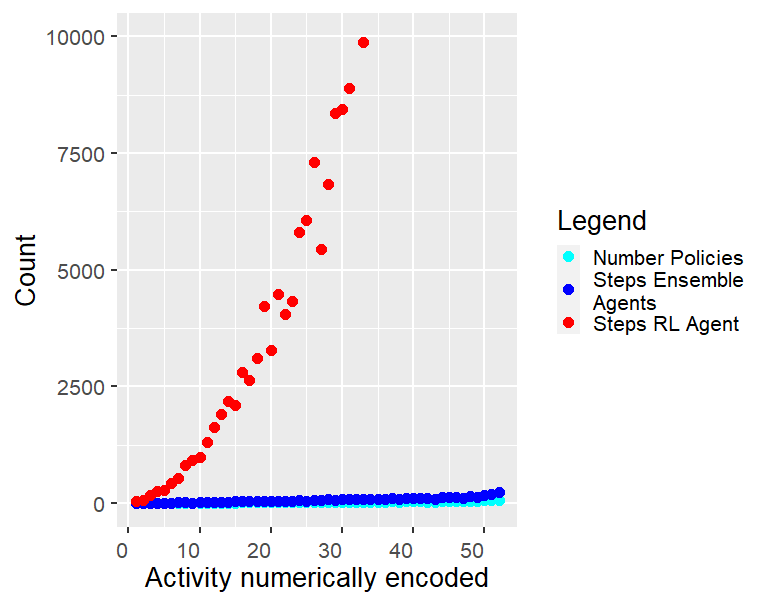}
            \caption{10 Episodes}
            \label{fig:mixed_10}
    \end{subfigure}
    \begin{subfigure}{.33\textwidth}
            \includegraphics[width=\textwidth]{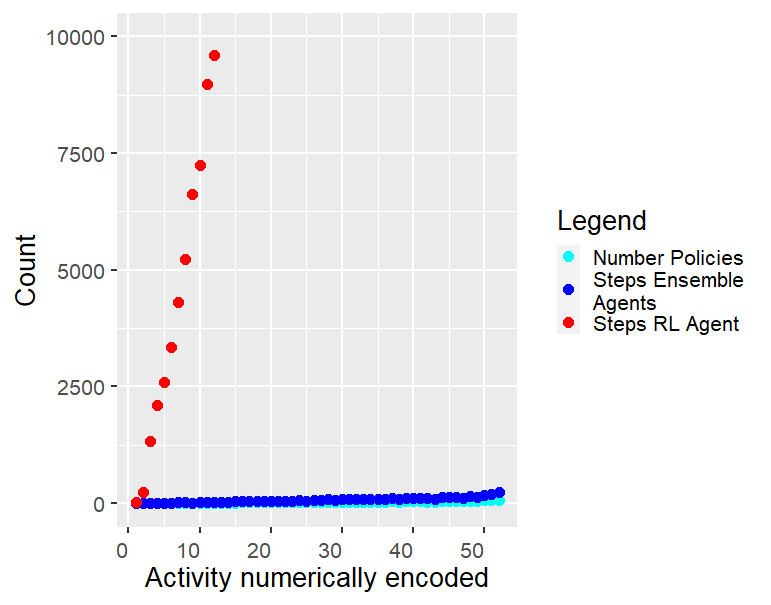}
            \caption{100 Episodes}
            \label{fig:mixed_100}
    \end{subfigure}
    \caption{Required steps (y-axis) of ensemble agents (blue dots) and RL agent (red dots) per activity (x-axis) until policies are composed or trained. The turquoise dots show the number of required policies among each activity. In these diagrams, as well, the y-axis was limited to a range between 0 and 10000 for reasons of clarity. Therefore, some data points are missing from the diagrams because they lie outside the range shown. However, the trend of the data points does not change but increases the more complex the tasks become. Furthermore, it is evident from the data points shown that the measured values, i.e. required execution steps, diverge strongly.}
    \label{fig:steps2Activity}
\end{figure*}

\noindent To show that the proposed policy composition approach outperforms the DQNN agent in terms of velocity, the steps executed were counted until correct, i.e. activity description compliant, policies could be provided for each activity. Fig.~\ref{fig:steps2Activity} shows that our presented approach needs significantly fewer steps than the DQNN agent to provide correct policies for each tested activity. \\

\begin{figure*}[!htbp]
    \centering
    \begin{subfigure}{.33\textwidth}
    \includegraphics[width=\textwidth]{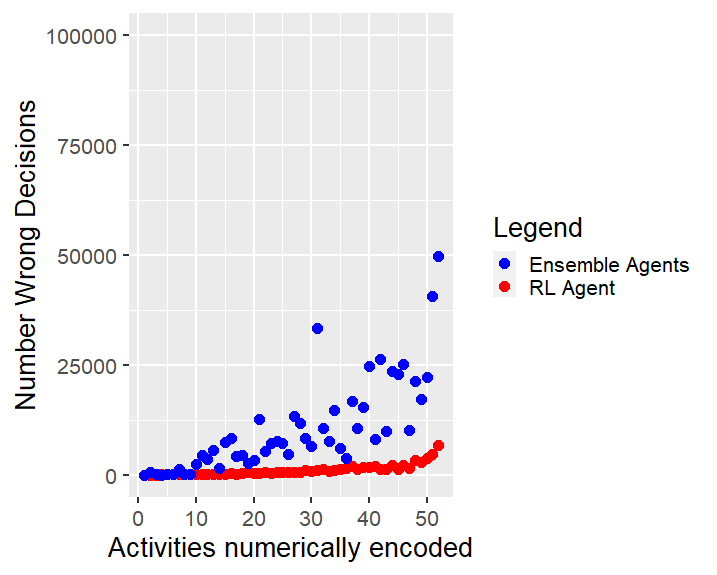}
    \caption{1 Episode}
    \label{fig:wrongDec_task_1}
    \end{subfigure}
    \begin{subfigure}{.33\textwidth}
    \includegraphics[width=\textwidth]{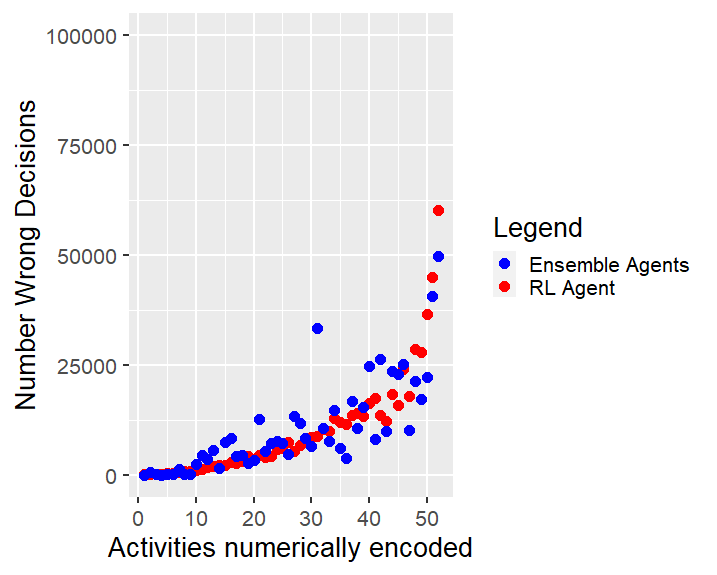}
    \caption{10 Episodes}
    \label{fig:wrongDec_task_10}
    \end{subfigure}
    \begin{subfigure}{.33\textwidth}
    \includegraphics[width=\textwidth]{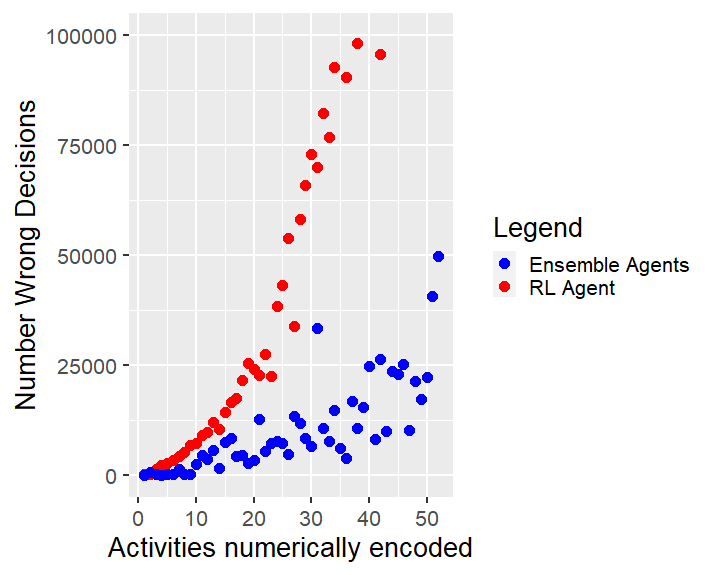}
    \caption{100 Episodes}
    \label{fig:wrongDec_task_100}
    \end{subfigure}
    \caption{Number of incorrect decisions (y-axis) made for each activity (x-axis) during a) policy composition by ensemble agents (blue dots), and b) policy training by RL agent (red dots).}
    \label{fig:wrongDecs2Acts}
\end{figure*}

\noindent We also counted the number of wrong, i.e. reward-minimising, actions that both approaches performed for each tested activity. Fig.~\ref{fig:wrongDecs2Acts} presents the corresponding results for each evaluation run. Fig.~\ref{fig:wrongDec_task_1} shows that the ensemble agents perform significantly more wrong actions. However, Fig.~\ref{fig:wrongDec_task_10} and \ref{fig:wrongDec_task_100} illustrate that the more episodes the RL agent has to perform, the more wrong actions it shows. Furthermore, it has to be taken into account that the ensemble agents considered all potentially possible actions, whereas the RL agent only knew the possible actions of the corresponding activity. Thus, the ensemble agents could also make significantly more mistakes, as they could choose actions from a much larger action space encompassing all activities. In addition, the simulation function also penalised actions of the ensemble agents that were not listed in the current activity description, although they had the same purpose as the intended actions. Furthermore, the RL agent did not provide reliable policies after 1 training episode. Thereby, the RL agent needed at least 100 episodes to provide correct policies for at least 14 activities with a maximum action sequence length of 14, while the proposed ensemble approach could provide correct policies for all tested 52 activities within 1 episode. \\

\noindent Considering the evaluation results of both approaches, we can conclude for \textbf{RQ2} that our approach is indeed able to speed up policy delivery in terms of fewer episodes (i.e. only 1 episode) and steps compared to agents using RL (i.e. DQNN).  \\

\begin{figure*}[!htbp]
    \centering
    \begin{subfigure}{.33\textwidth}
    \includegraphics[width=\textwidth]{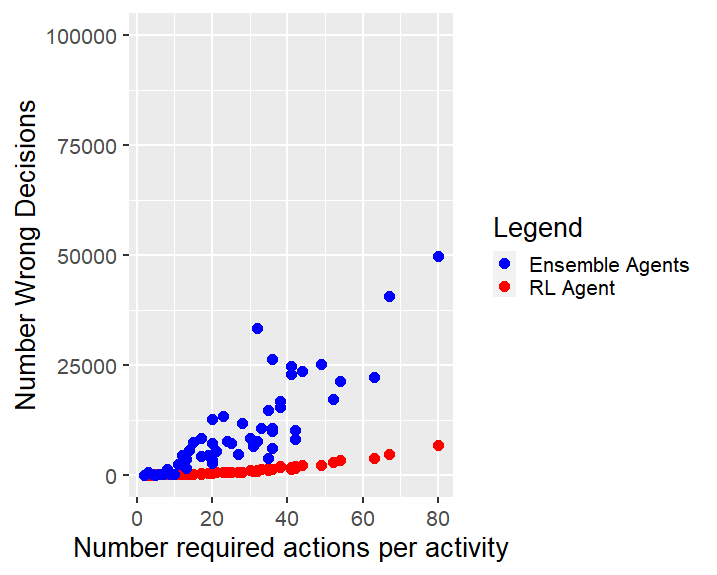}
    \caption{1 Episode}
    \label{fig:wrongDec_NumPolicies_1}
    \end{subfigure}
    \begin{subfigure}{.33\textwidth}
    \includegraphics[width=\textwidth]{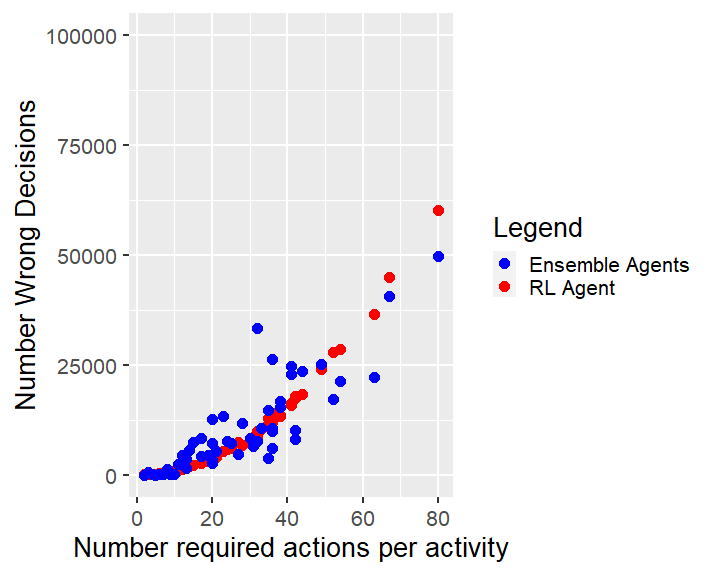}
    \caption{10 Episodes}
    \label{fig:wrongDec_NumPolicies_10}
    \end{subfigure}
    \begin{subfigure}{.33\textwidth}
    \includegraphics[width=\textwidth]{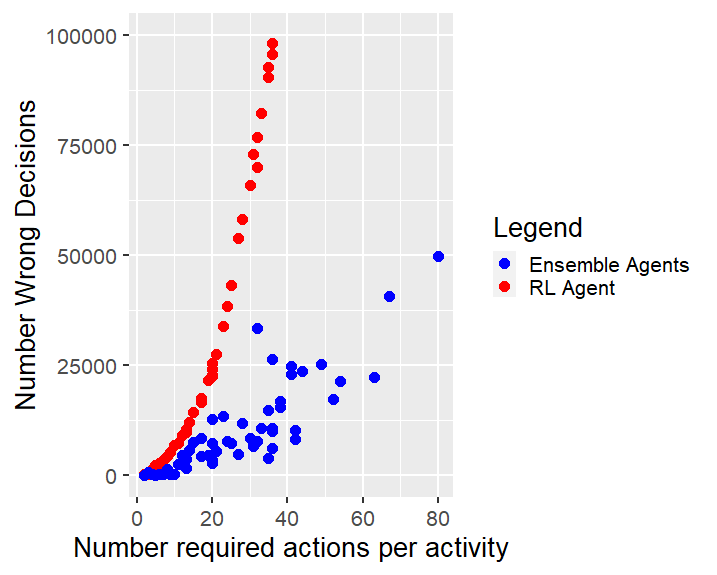}
    \caption{100 Episodes}
    \label{fig:wrongDec_NumPolicies_100}
    \end{subfigure}
    \caption{Number of incorrect decisions (y-axis) w.r.t. the number of required policies per activity (x-axis) made by Ensemble agents (blue dots) during policies composition and RL agent (red dots) during policies training.}
    \label{fig:wrongDecs_NumPolicies}
\end{figure*}

\subsection{Limitations in the Evaluation}
\noindent In the experiments conducted, only one activity dataset, namely the \emph{Virtual Home} dataset, and only one RL algorithm, i.e. DQNN, were evaluated and compared with the proposed approach. If more time had been available, more different RL algorithms and activity datasets would normally have had to be evaluated to show the generalisability of the presented approach for other domains as well. However, the \emph{Virtual Home} dataset offers many different activities in domestic environments or contexts, so that different contexts and activities can be evaluated using the dataset. Furthermore, the developed MDP ontology generalises and abstracts activities and tasks, suggesting that the evaluated approach is likely to be transferable and generalisable to further datasets, while no serious difference in learning rates between the different RL algorithms and the deep q-net algorithm is expected. Furthermore, we anticipate that the other RL algorithms are likely to have similar performance measures among themselves due to their iterative approach and depending on the size of the state space, but this could not be tested during the evaluation.

\section{Conclusion}
\label{sec:conclusion}
\noindent In this paper, we have presented a simulation-based approach that uses knowledge graphs to describe activities, agent ensembles and embeddings of MDP entities for contextual policy composition. The idea and goal of the approach is to support computational agents in heterogeneous environments and contexts by providing policies, i.e., sequences of actions on demand, that can be used in a variety of contexts. The approach aims at offering alternative strategies to agents so that they can choose the ones that are viable for them. In various application fields, e.g., service robotics, it can be observed that agents have to cope in heterogeneous environments and arrange with rapidly changing contexts. Moreover, new capabilities and actions of the agent may be required, especially when the agent is presented with new tasks in an environment at runtime. The proposed approach attempts to meet the aforementioned requirements and, as the evaluation results indicate, offers a fast-functioning alternative to RL. \\

\noindent For the conducted evaluation, we adopted the \emph{Virtual Home (VH)} dataset as ground truth for domestic activities and compared the policy delivery speed of a DQNN agent with the speed of our approach. We were able to show that our approach was able to provide proven policies for all tested domestic activities consistently to the VH activity descriptions, within only one episode and with significantly fewer search steps than the DQNN agent required. The related work considered has shown that RL approaches are widely used by computational agents. However, they have the disadvantage that, depending on the size of the state and action space of MDP activities, time-consuming training procedures are required until reward-maximising and goal-fulfilling policies can be learned that enable an agent to successfully perform its activities on demand. Our approach does not require guided training procedures, as it uses entity embeddings, trained continuously from either activity-specific datasets or MDP knowledge graphs that either are derived from recorded time-series datasets or created by domain experts. Furthermore, our proposed approach has been shown to be able to significantly constrain the state and action space, enabling the proposed platform to predict the most likely neighbouring states and corresponding actions that may occur in the agent's context. \\

\noindent The proposed approach enables agents to act across contexts in a predictive manner and thus arrive at decisions more quickly. Moreover, the approach presented addresses policy requirements of agents and allows them to perform activities without having to train policies beforehand. To make our obtained results replicable, our generated semantic activity descriptions, trained embedding vectors, and source code are freely and openly available on Github\footnote{\url{https://github.com/nmerkle/SW_Journal}} and Figshare\footnote{\url{https://figshare.com/s/ad977340ac008ef22ada}}. The source code is implemented with JavaScript and NodeJS and can be executed on all common operating systems, as both JavaScript and NodeJS\footnote{\url{https://nodejs.org/de}} are platform-independent. \\

\noindent Future work will focus on the integration and evaluation of heterogeneous datasets and agent activities from domains other than the considered one. In addition, we will consider generative models, such as \emph{Transformer} models to allow different simulations of contexts and activities that also take into account exceptional environmental events. The aim is to arrive at generalisable policies in this way that also apply to random situations and contexts.





\nocite{*}
\bibliographystyle{ios1}           
\bibliography{bibliography}        

%

\end{document}